\documentclass{article}

\PassOptionsToPackage{numbers, square}{natbib}
\usepackage[final]{neurips_2024}

\usepackage[utf8]{inputenc} 
\usepackage[T1]{fontenc}    
\usepackage{hyperref}       
\usepackage{url}            
\usepackage{booktabs}       
\usepackage{amsfonts}       
\usepackage{nicefrac}       
\usepackage{microtype}      
\usepackage[table]{xcolor}         
\usepackage{amsmath,amsfonts}
\usepackage{amssymb}
\usepackage{amsthm}
\usepackage[ruled,commentsnumbered]{algorithm2e}
\SetKwRepeat{Do}{repeat}{until}
\usepackage{array}
\usepackage{bm}
\usepackage{booktabs}
\usepackage{color}
\usepackage{soul}
\usepackage{ifthen}
\usepackage{longtable}
\usepackage{mathtools}
\usepackage{makecell}
\usepackage{multirow}
\usepackage{nomencl}
\usepackage[short]{optidef}
\usepackage{pifont}
\usepackage{tabularx,ragged2e}
\usepackage{threeparttable}
\newcolumntype{C}{>{\Centering\arraybackslash}X} 
\newcolumntype{P}[1]{>{\centering\arraybackslash}p{#1}}
\newcolumntype{M}[1]{>{\centering\arraybackslash}m{#1}}

\newtheorem{remark}{Remark}
\newtheorem{corollary}{Corollary}

\title{GLinSAT: The General Linear Satisfiability Neural Network Layer By Accelerated Gradient Descent}

\author{%
	Hongtai Zeng$^{1}$ \quad Chao Yang$^{2}$ \quad Yanzhen Zhou$^1$ \quad Cheng Yang$^2$ \quad Qinglai Guo$^1$\thanks{Corresponding author.} \\
	$^1$ State Key Laboratory of Power Systems, Department of\\
	Electrical Engineering, Tsinghua University \\
	$^2$ Decision Intelligence Lab, Alibaba DAMO Academy \\
	\texttt{zenght20@mails.tsinghua.edu.cn, xiuxin.yc@alibaba-inc.com, zhouyzh@126.com,} \\	
	\texttt{charis.yangc@alibaba-inc.com, guoqinglai@tsinghua.edu.cn}
}

\begin{document}

\maketitle

\begin{abstract}
Ensuring that the outputs of neural networks satisfy specific constraints is crucial for applying neural networks to real-life decision-making problems. In this paper, we consider making a batch of neural network outputs satisfy bounded and general linear constraints. We first reformulate the neural network output projection problem as an entropy-regularized linear programming problem. We show that such a problem can be equivalently transformed into an unconstrained convex optimization problem with Lipschitz continuous gradient according to the duality theorem. Then, based on an accelerated gradient descent algorithm with numerical performance enhancement, we present our architecture, GLinSAT, to solve the problem. To the best of our knowledge, this is the first general linear satisfiability layer in which all the operations are differentiable and matrix-factorization-free. Despite the fact that we can explicitly perform backpropagation based on automatic differentiation mechanism, we also provide an alternative approach in GLinSAT to calculate the derivatives based on implicit differentiation of the optimality condition. Experimental results on constrained traveling salesman problems, partial graph matching with outliers, predictive portfolio allocation and power system unit commitment demonstrate the advantages of GLinSAT over existing satisfiability layers. Our implementation is available at {\url{https://github.com/HunterTracer/GLinSAT}}.

\end{abstract}

\section{Introduction}
Constrained decision-making problems are pervasive across various disciplines. For example, logistics companies need to arrange delivery routes to minimize transportation costs while ensuring that all orders are delivered on time. Power system operators need to decide how to allocate electricity production between different power plants to meet ever-changing electricity demand while maintaining system stability. Unfortunately, directly solving these complex constrained decision-making problems via commercial optimization solvers requires a large amount of time. As a result, in scenarios that require rapid response, traditional solvers may not be suitable due to their long computation time. With the development of deep learning, it is hopeful that neural networks can capture the domain characteristics and complex relationships involved in constrained decision-making problems through their powerful expressive capability and the solution time can be thus reduced. In recent years, research on how to use neural networks to solve constrained decision-making problems has become a topic of general interest. Despite the great success of neural networks on classification and regression tasks, making the outputs of neural networks satisfy specific constraints is not straightforward, which still needs to be further investigated.

A natural idea to impose constraints on the neural network outputs is to penalize the constraint violation in the training stage of supervised learning or reinforcement learning {\cite{karalias2020erdos, bohez2019value, bhatia2019resource}}. However, such an approach requires a careful selection of each penalty coefficient to achieve a balance between decision objectives and constraint violations. As the complexity of constraints increases, choosing appropriate penalty factors may require a large number of attempts, which is time-consuming. Moreover, it is often difficult to theoretically guarantee the boundedness of constraint violations {\cite{wang2022towards}}, which makes penalty-based methods less attractive. Ref. {\cite{zhao2023ensuring}} managed to determine the width of neural networks required for ensuring feasibility by modeling these networks using binary variables and solving a complex mixed-integer bilevel programming. However, this approach necessitates shrinking the original feasible region and can only handle inequality constraints, which limits its broader application. There are also methods in the literature that are better suited for inequality constraint satisfaction. Ref. {\cite{tabas2022computationally}} uses gauge function to map the neural network outputs from a $\ell^\infty$-norm unit ball into a given polyhedral. Despite its success in the field of control, this method may encounter difficulties in handling equality constraints since the polyhedral need to contain the origin as an interior point. Ref. {\cite{tordesillas2023rayen}} first calculates a reference point within the interior of a convex region using convex programming in the offline stage, and then computes a feasible point based on this reference point through simple arithmetic operations in the online stage. Despite its efficiency potentially being affected when constraints are not fixed, the method may encounter difficulties in satisfying equality constraints, as it requires computing the null space of the equality constraints. Since the basis for the null space of a matrix is typically dense, calculating the null space for large matrices may present both efficiency and memory challenges. Another way to encode constraints in neural networks is to reformulate the original problem as a Markov decision problem {\cite{bello2016neural, khalil2017learning}}. During the solution process, the decision variables are generated one by one and the value range of the next variable is determined by the value of the current variable so that constraints can be satisfied compulsorily. However, not all decision-making problems can be equivalently converted to Markov decision problems which limits the application of such an approach.

Due to the limitations of the above approaches, many researchers want to use a more reliable way to ensure that the outputs of neural networks satisfy specific constraints. A promising way is to integrate optimization solvers as neural network layers. 
When we embed a solver for end-to-end learning, we need to pay special attention to the following two issues: the first one is the \textit{\textbf{supported constraint types}}, and the second one is the \textit{\textbf{efficiency}}.

As for the issue of supported constraint types, some frameworks can directly impose combinatorial constraints on neural network outputs through integrating black-box commercial mixed-integer programming solvers at the cost of inexact gradient estimates and poor utilization of GPUs (since modern commercial solvers are CPU-based) {\cite{poganvcic2019differentiation, berthet2020learning, paulus2021comboptnet}}. These approaches need to solve combinatorial optimization problems in both training and inference stages, which is time-consuming. Instead of directly handling the combinatorial constraints, some researchers manage to make neural network outputs satisfy constraints obtained from the continuous relaxation of the original problem, e.g. the widely used double stochastic matrix constraint in Ref. {\cite{santa2017deeppermnet, wang2019learning, cuturi2019differentiable}} solved by Sinkhorn algorithm {\cite{sinkhorn1967concerning, cuturi2013sinkhorn, chakrabarty2021better}}. Another example is the positive semi-definite (PSD) matrix constraint with unit diagonals as a continuous relaxation of the original MAXSAT problem  {\cite{wang2019satnet}}. However, both of the above methods can only express specific constraints, which limits their application. Recently, LinSAT, which is based on a generalized Sinkhorn algorithm, is proposed to impose positive linear constraints on neural network outputs {\cite{wang2023linsatnet}}. However, the requirement for all elements in constraints to be non-negative limits the application of LinSAT. For example, even for a simple constraint $x \le y$, namely $x - y \le 0$, LinSAT cannot be used due to the negative coefficient in front of $y$, which shows the limited expressiveness of LinSAT. Decision variables with negative constraint coefficients occur a lot in real-life decision-making problems, such as the bin packing problem {\cite{mezghani2023evolution}}, chemical process scheduling {\cite{floudas2005mixed}}, power system unit commitment {\cite{knueven2020mixed}}, etc.

To deal with general linear constraints in a differentiable way, currently, there are two main approaches, CvxpyLayers {\cite{agrawal2019differentiable}} and OptNet {\cite{amos2017optnet}}. However, when solving a batch of problems, both of them may encounter efficiency issues. Although CvxpyLayers can achieve parallelism through multiprocessing on the CPU, there are only a dozen of cores in one CPU, leading to limited parallelism performance. On the other hand, OptNet presents a GPU-based batch quadratic programming interior point solver where batch matrix factorization are performed to accelerate the solution process. Unfortunately, batch matrix factorization may be still a computational bottleneck even when GPU is used. Although some scholars have also studied how to parallelize parts of operations in matrix factorization on the GPU {\cite{rennich2016accelerating, tang2017multithreaded}}, the degree of parallelism still highly depend on the structure of the matrix and its elimination tree. Two nodes in the elimination tree can be computed in parallel only when there is no direct branch connecting them. As a result, matrix factorization cannot fully utilize the parallel computing ability of the GPU due to the sequential characteristics in the elimination tree {\cite{davis2016survey}}.

In this paper, we investigate how to apply bounded and general linear constraints to neural network outputs in a differentiable way while ensuring batch processing can be performed efficiently on the GPU.

\textbf{The contributions of this paper include:}

1) To impose general linear and bounded constraints on neural network outputs in a differentiable way, we formulate the corresponding projection problem as an entropy-regularized linear programming where negative logistic entropic regularization terms are added into the objective function. We show that such an entropy-regularized linear programming problem can be transformed into an unconstrained convex optimization problem with Lipschitz continuous gradient, and thus can be solved by gradient descent based algorithms where no matrix factorization operation is required.

2) We design GLinSAT, a general linear satisfiability layer to impose linear constraints on neural network outputs based on a state-of-the-art accelerated gradient descent algorithm with numerical performance enhancement. Since the main operations in GLinSAT is matrix-vector product and no matrix factorization is involved, it is convenient to execute these operations in parallel on the GPU. Although all the operations involved in GLinSAT are differentiable which means that we can directly use the automatic differentiation mechanism to perform back propagation, we also provide an alternative way for derivative calculation based on the optimality condition to reduce the memory consumption.

\begin{figure}[h]
	\centering
	\includegraphics[width=\textwidth]{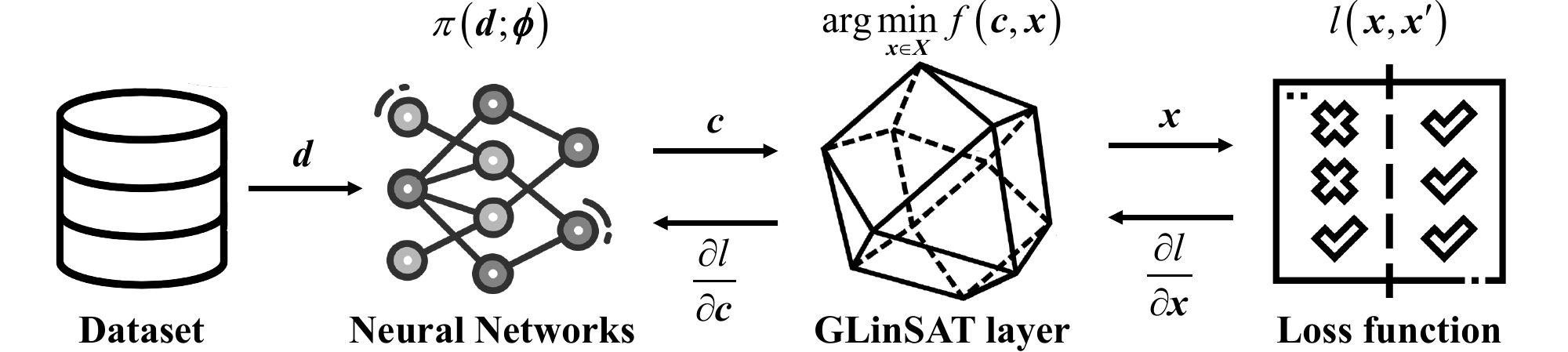}
	\caption{A pipeline that shows how GLinSAT layer works.}
	\label{fig1}
\end{figure}

3) We then provide experimental results to demonstrate the capabilities of our proposed method. Experiments on constrained traveling salesman problems, partial graph matching with outliers, predictive portfolio allocation and power system unit commitment show the efficacy of our proposed method and advantages over existing methods. A comparison of methodologies for imposing constraints on neural networks outputs is presented in Table {\ref{tab_contribution}}. A pipeline that shows how our approach works is provided in Fig. {\ref{fig1}}.
\begingroup
\renewcommand{\arraystretch}{1.1}
\begin{table*}[h]
	\centering
	\caption{Comparison with existing optimizer layers for imposing constraints on the outputs of neural networks}
	\begin{tabularx}{\textwidth}{M{0.07\textwidth}M{0.145\textwidth}M{0.12\textwidth}M{0.095\textwidth}M{0.095\textwidth}M{0.06\textwidth}M{0.095\textwidth}M{0.095\textwidth}}
		\toprule
		Ref. & Abbr. of approach & Constraint type & GPU parallel computing & Matrix factorization free & Exact gradient & Explicit backpropagation & Implicit backpropagation \\
		\midrule
		\cite{poganvcic2019differentiation, berthet2020learning, paulus2021comboptnet} & Perturbed optimizer & \makecell[b]{combina-\\torial} & \textbf{\---} & \textbf{\---} & \ding{55} & \ding{55} & \ding{51} \\
		\cite{santa2017deeppermnet, wang2019learning, cuturi2019differentiable} & Sinkhorn & double stochastic matrix & \ding{51} & \ding{51} & \ding{51} & \ding{51} & \ding{55} \\
		\cite{wang2019satnet} & SATNet & PSD matrix with unit diagonals & \ding{51} & \ding{51} & \ding{51} & \ding{55} & \ding{51} \\
		\cite{wang2023linsatnet} & LinSAT & positive linear & \ding{51} & \ding{51} & \ding{51} & \ding{51} & \ding{55} \\
		\cite{agrawal2019differentiable} & CvxpyLayers & linear and cone & \ding{55} & \ding{55} & \ding{51} & \ding{55} & \ding{51} \\
		\cite{amos2017optnet} & OptNet & linear & \ding{51} & \ding{55} & \ding{51} & \ding{55} & \ding{51} \\
		\midrule
		This article & GLinSAT & linear & \ding{51} & \ding{51} & \ding{51} & \ding{51} & \ding{51} \\
		\bottomrule
	\end{tabularx}
	\label{tab_contribution}%
	\begin{tablenotes}
		\item[a] Note: Explicit/Implicit backpropagation means that this algorithm performs backward propagation based on automatic differentiation mechanism/implicit differentiation. \textbf{\---} means that this algorithm feature is dependent on the implementation of the backend solver.
	\end{tablenotes}
\end{table*}%
\endgroup

\section{Methodology}
Sec. {\ref{sec3.1}} formulates the neural network output projection problem as an entropy-regularized linear programming by introducing logistic entropy regularization terms in the objective function. Based on duality theorem, the original problem can be transformed into an unconstrained convex optimization problem with Lipschitz continuous gradient. Then, in Sec. {\ref{sec3.2}}, based on a variant of accelerated gradient descent method, we design GLinSAT, which solves the projection problem using a GPU-friendly algorithm with several numerical enhancements. The corresponding time complexity is also provided in Sec. {\ref{sec3.2}}. Moreover, although all the operations in the forward pass of GLinSAT are differentiable, in Sec. {\ref{sec3.3}}, we provide an alternative way based on the optimality condition to calculate the derivatives for memory saving.

\subsection{Reformulation of the neural network output projection problem} \label{sec3.1}
Here, we want to use a differentiable way to project the output of the neural network $\bm{c}' \in \mathbb{R}^{n'}$ into variables $\bm{x}' \in \mathbb{R}^{n'}$ that are as similar as possible but satisfy the following constraints ({\ref{cons_orig}}).
\begin{subequations}
	\label{cons_orig}
	\begin{gather}
		\bm{A}_1' \bm{x}' \leq \bm{b}_1' \\
		\bm{A}_2' \bm{x}' \geq \bm{b}_2' \\
		\bm{A}_3' \bm{x}' = \bm{b}_3' \\
		\bm{l}' \leq \bm{x}' \leq \bm{u}'
	\end{gather}
\end{subequations}
where $\bm{A}_1' \in \mathbb{R}^{m_1' \times n'}$, $\bm{A}_2' \in \mathbb{R}^{m_2' \times n'}$, $\bm{A}_3' \in \mathbb{R}^{m_3' \times n'}$, $\bm{b}_1' \in \mathbb{R}^{m_1'}$, $\bm{b}_2' \in \mathbb{R}^{m_2'}$, $\bm{b}_3' \in \mathbb{R}^{m_3'}$, $\bm{l}', \bm{u}' \in \mathbb{R}^{n'}$. Moreover, we also suppose that the feasible region in ({\ref{cons_orig}}) is non-empty. 

Apparently, any general linear constraints with bounded variables like ({\ref{cons_orig}}) can be converted to the standard form like ({\ref{cons_standard}}) by shifting bounds and introducing slack variables (see Appendix {\ref{appendix.a1}}):
\begin{subequations}
	\label{cons_standard}
	\begin{gather}
		\bm{A} \bm{x} = \bm{b} \\
		\bm{0} \leq \bm{x} \leq \bm{u}
	\end{gather}
\end{subequations}
where $m = m'_1 + m'_2 + m'_3$, $n = n' + m'_1 + m'_2$, $\bm{A} \in \mathbb{R}^{m \times n}$, $\bm{b} \in \mathbb{R}^m$, $\bm{u} \in \mathbb{R}_+^n$. Here, we denote the vector obtained from padding $\left(m'_1 + m'_2\right)$ zeros after the original vector $\bm{c}'$ as $\bm{c}$. Now, the original problem is transformed into a problem of projecting $\bm{c} \in \mathbb{R}^n$ onto $\bm{x} \in \mathbb{R}^n$ that satisfy constraints ({\ref{cons_standard}}). In the following sections, we mainly focus on such a transformed problem in standard form.

In this paper, we aim for the vector $\bm{x}$ after projection to be as close as possible to the vector $\bm{c}$ prior to projection, while adhering to specified constraints. Here, we use the dot product as a measure of vector similarity. Consequently, our objective function becomes that of maximizing the dot product of vectors $\bm{c}$ and $\bm{x}$, which is equivalently described as minimizing the dot product of $-\bm{c}$ and $\bm{x}$. Besides, as pointed by {\cite{wilder2019melding}}, the optimal solution to an linear programming may not be differentiable (or even continuous) with respect to its parameters. Therefore, additional regularization terms need to be included in the objective to make the optimization problem differentiable. Inspired by entropy-regularized optimal transport, here we formulate the projection problem as an entropy-regularized linear programming to make the entire problem differentiable. Logistic entropy regularization terms are added into the objective as follows:
\begin{subequations}
	\label{prob_primal}
	\begin{gather}
		\mathop {\min }\limits_{\bm{0} \leq \bm{x} \leq \bm{u}}  f\left(\bm{x}\right) = \mathop {\min }\limits_{\bm{0} \leq \bm{x} \leq \bm{u}} -\bm{c}^T \bm{x} + \frac{1}{\theta }\sum\limits_{j = 1}^n {\left( {\frac{{{x_j}}}{{{u_j}}}\log \frac{{{x_j}}}{{{u_j}}} + \left( {1 - \frac{{{x_j}}}{{{u_j}}}} \right)\log \left( {1 - \frac{{{x_j}}}{{{u_j}}}} \right)} \right)}  \\
		\label{cons_eq}
		{\rm{s.t.}}\; \bm{A} \bm{x} = \bm{b}
	\end{gather}
\end{subequations}
where $\theta > 0$ is the inverse temperature parameter that controls the approximation degree between the entropy-regularized problem and the original linear programming. The regularization coefficient $1/\theta$
controls the smoothness of the outputs. The smaller $1/\theta$ is, the more the outputs tend to be at the extreme point of the feasible region. As $\theta \to + \infty$, the optimal solution of the entropy-regularized problem should approach that of the original linear programming. 

\begin{remark}
	It is noteworthy that unlike entropy-regularized optimal transport problems where only regularization terms in the form of $x \log x$ are involved in the objective, here logistic entropy regularization terms with respect to both $x / u$ and its complement $1 - x / u$ are added into the objective. \textbf{Actually, additionally incorporating the complementary entropy regularization terms is the most important part for the derivation of the Lagrange dual problem.} Otherwise, we cannot obtain a simple closed-form expression of the dual objective in the following derivation. Similarly, if we use the common $\ell^2$-norm as the regularization term, we cannot obtain an analytical expression either.
\end{remark}

If we denote the dual variables with respect to the equality constraints ({\ref{cons_eq}}) as $\bm{y}$, the Lagrange dual function for ({\ref{prob_primal}}) can be expressed as follows:
\begin{equation}
	\label{lagrange_dual}
	g\left( {\bm{y}} \right) = \mathop {\inf }\limits_{\bm{0} \leq \bm{x} \leq \bm{u}} \left( -\bm{c}^T \bm{x} + \frac{1}{\theta } \bm{1}^T \left(\frac{\bm{x}}{\bm{u}} \circ \bm{\log} \frac{\bm{x}}{\bm{u}} + \left(\bm{1} - \frac{\bm{x}}{\bm{u}}\right) \circ \bm{\log} \left(\bm{1} - \frac{\bm{x}}{\bm{u}}\right)\right) - \bm{y}^T\bm{A} \bm{x}  \right) + {\bm{b}^T}\bm{y}
\end{equation}
where $\bm{a} \circ \bm{b}, \frac{\bm{b}}{\bm{a}}$ represents the element-wise multiplication and division of vector $\bm{a}$ and $\bm{b}$ respectively.

Since the derivative magnitude of $x \log x + \left(1 - x\right) \log \left(1 - x\right)$ tends to infinity when $x \to 0^+$ or $x \to 1^-$, the infimum in ({\ref{lagrange_dual}}) can be attained only on a stationary
point instead of a boundary point. When the infimum in ({\ref{lagrange_dual}}) is attained, by making the derivative of the inner function equal to zero, we have:
\begin{equation}
	\label{optimal_lagrange_dual}
	-\bm{c} - \bm{A}^T \bm{y} + \frac{\bm{1}}{\theta \bm{u}} \circ \bm{\log} \frac{\bm{x}}{{\bm{u} - \bm{x}}} = \bm{0}
\end{equation}

After simplifying the above formula, we can get that when the infimum in ({\ref{lagrange_dual}}) is attained, the optimal value of $\bm{x}\left(\bm{y}\right)$ can be expressed as:
\begin{equation}
	\label{optimal_x}
	\bm{x}\left(\bm{y}\right) = \bm{u} \circ \bm{\sigma} \left( { - {\theta}\bm{u} \circ \left(- \bm{c} - \bm{A}^T \bm{y}\right)} \right)
\end{equation}
where $\bm{\sigma}\left(\cdot\right)$ is the sigmoid function.

Substituting equation ({\ref{optimal_x}}) into equation ({\ref{lagrange_dual}}), we have:
\begin{equation}
	g\left( {\bm{y}} \right) = \frac{1}{{{\theta }}} \bm{1}^T {\bm{\log} {\bm{\sigma} \left( {{\theta }\bm{u} \circ \left(-\bm{c} - \bm{A}^T\bm{y}\right)} \right)}} + {\bm{b}^T}\bm{y}
\end{equation}

Since $\bm{\log} \bm{\sigma} \left(\cdot\right)$ is a strictly concave function, by minimizing the opposite of $g\left( {\bm{y}} \right)$, we can obtain the following Lagrange dual problem ({\ref{prob_dual}}), which is exactly an unconstrained convex optimization problem.
\begin{equation}
	\label{prob_dual}
	\mathop {\min }\limits_{\bm{y} \in \mathbb{R}^m} - g\left(\bm{y}\right) = \mathop {\min }\limits_{\bm{y} \in \mathbb{R}^m} - \frac{1}{{{\theta }}} \bm{1}^T {\bm{\log} {\bm{\sigma} \left( {{\theta }\bm{u} \circ \left(-\bm{c} - \bm{A}^T\bm{y}\right)} \right)}} - {\bm{b}^T}\bm{y}
\end{equation}

We can easily show that $f\left(\bm{x}\right)$ is a strongly convex function and $-g\left(\bm{y}\right)$ has Lipschitz continuous gradient (see explanations in Appendix {\ref{appendix.a2}}). Therefore, gradient descent based algorithms can be directly applied to solve such a problem.

\subsection{Forward pass in GLinSAT} \label{sec3.2}
In the previous section, we have shown that the original entropy-regularized linear programming problem ({\ref{prob_primal}}) can be equivalently converted into an unconstrained convex optimization problem ({\ref{prob_dual}}) with Lipschitz continuous gradient. Theoretically, it can be solved readily through gradient descent based method. However, in the actual calculation process, it will be hard to choose a suitable step size if we just use vanilla gradient descent method. If the step size is much greater than the local Lipschitz constant, the algorithm may diverge. Otherwise, the convergence may be too slow.

Considering the strong convexity property of the entropy regularization terms, here we use a variant of accelerated gradient descent method, adaptive primal-dual accelerated gradient descent (APDAGD), which can adaptively approximate the local Lipschitz constant {\cite{dvurechensky2018computational}}. The detailed procedure of solving the entropy-regularized linear programming problem ({\ref{prob_primal}}) in GLinSAT is provided in Algorithm {\ref{alg_apdagd}}.

Compared with the original version of APDAGD, here we improve the numerical performance of Algorithm {\ref{alg_apdagd}} from the following two aspects. First, we use a smoother way to update the approximation of the local Lipschitz constant $M$ in GLinSAT. In Algorithm {\ref{alg_apdagd}}, $M$ is decreased only when the decrease of the dual objective satisfies the corresponding condition for at least two consecutive times. As a result, when $M$ is already a good estimate of the local Lipschitz constant, the frequency of needless updates can be reduced, which will lead to less computation time. Second, to handle the round-off error, we also use a small number $\delta$ to relax the criterion for the decrease of the objective function. Otherwise, due to the existence of numerical error, the criterion of sufficient decrease in objective may be never satisfied. If we do not relax the criterion, $M$ may become a large number and the algorithm will get stuck.

In addition, it is noteworthy that most of the operations involved in Algorithm {\ref{alg_apdagd}} are calculation of matrix-vector products, vector-vector element-wise products and unary functions. Therefore, it is convenient to execute these operations in parallel on the GPU for solving a batch of entropy-regularized linear programming problems.

As for the time complexity of Algorithm {\ref{alg_apdagd}}, based on Theorem 1 and Theorem 2 in the supplementary material of {\cite{dvurechensky2018computational}}, it can be easily proved that the number of iterations required by Algorithm 1 is roughly proportional to $\sqrt{\theta}$ and inversely proportional to $\sqrt{\varepsilon}$. The corresponding result is given in Corollary {\ref{corollary_1}} and the detailed discussions can be found in Appendix {\ref{appendix.a3}}.

\begin{algorithm}[H]
	\SetAlgoLined
	\SetKwFunction{True}{True}
	\SetKwFunction{False}{False}
	\caption{Solving the entropy-regularized linear programming problem in GLinSAT}\label{alg_apdagd}
	\KwIn{$\bm{A} \in \mathbb{R}^{m \times n}$, $\bm{b} \in \mathbb{R}^{m}$, $\bm{c} \in \mathbb{R}^{n}$, $\bm{u} \in \mathbb{R}_+^{n}$, inverse temerature $\theta > 0$, tolerance $\varepsilon > 0$, initial estimate of Lipschitz constant $L^{\left(0\right)}$, initial estimate of dual variables $\bm{y}^{\left(0\right)}$, numerical precision $\delta > 0$}
	Set $k = 0$, $M^{\left(0\right)} = L^{\left(0\right)}$, $\bm{\eta}^{\left(0\right)} = \bm{\zeta}^{\left(0\right)} = \bm{y}^{\left(0\right)}$, $\bm{{x}}^{\left(0\right)} = \bm{u} \circ \bm{\sigma} \left(- \theta \bm{u} \circ \left(- \bm{c} - \bm{A}^T \bm{y}^{\left(0\right)}\right)\right)$, $\beta^{\left(0\right)} = \alpha^{\left(0\right)} = 0$, $f$ = \False\;
	\While{${\left\| {\bm{A}\bm{x}^{\left(k\right)} - \bm{b}} \right\|_2} > \varepsilon$}{
		Set $\alpha^{\left(k + 1\right)} = \left(1 + \sqrt{1 + 4M^{\left(k\right)}\beta^{\left(k\right)}}\right) / \left(2M^{\left(k\right)}\right)$\;
		Set $\beta^{\left(k + 1\right)} = \beta^{\left(k\right)} + \alpha^{\left(k + 1\right)}$\;
		Set $\tau^{\left(k + 1\right)} = \alpha^{\left(k + 1\right)} / \beta^{\left(k + 1\right)}$\;
		Set $\bm{\lambda}^{\left(k + 1\right)} = \bm{\eta}^{\left(k\right)} + \tau^{\left(k + 1\right)} \left(\bm{\zeta}^{\left(k\right)} - \bm{\eta}^{\left(k\right)}\right)$\;
		Set $\bm{x}\left(\bm{\lambda}^{\left(k+1\right)}\right) = \bm{u} \circ \bm{\sigma} \left(- \theta \bm{u} \circ \left(- \bm{c} - \bm{A}^T \bm{\lambda}^{\left(k+1\right)}\right)\right)$\;
		Set $\bm{\zeta}^{\left(k + 1\right)} = \bm{\zeta}^{\left(k\right)} - \alpha^{\left(k + 1\right)} \left(\bm{A} \bm{x}\left(\bm{\lambda}^{\left(k+1\right)}\right) - \bm{b}\right)$\;
		Set $\bm{\eta}^{\left(k + 1\right)} = \bm{\eta}^{\left(k\right)} + \tau^{\left(k + 1\right)} \left(\bm{\zeta}^{\left(k + 1\right)} - \bm{\eta}^{\left(k\right)}\right)$\;
		\eIf{$\left(-g\left(\bm{\eta}^{\left(k+1\right)}\right)\right) - \left(-g\left(\bm{\lambda}^{\left(k+1\right)}\right)\right) - \delta \leq -\left\| \bm{A} \bm{x}\left(\bm{\lambda}^{\left(k+1\right)}\right) - b \right\|_2^2 / \left(2 M^{\left(k\right)}\right) $}{
			\eIf{$f$ = \True}{
				Set $M^{\left(k+1\right)} = M^{\left(k\right)} / 2$\;
			}{
				Set $M^{\left(k+1\right)} = M^{\left(k\right)}$\;
			}
			Set $\bm{{x}}^{\left(k+1\right)} = \bm{{x}}^{\left(k\right)} + \tau^{\left(k+1\right)} \left(\bm{x}\left(\bm{\lambda}^{\left(k+1\right)}\right) - \bm{{x}}^{\left(k\right)}\right)$, $f$ = \True\;
			Set $k = k+1$\;
		} {
			Set $M^{\left(k\right)} = 2M^{\left(k\right)}$, $f$ = \False\;
		}
	}
	\KwOut{Optimal primal variables $\bm{x}^{\left(k\right)}$, Optimal dual variables $\bm{\eta}^{\left(k\right)}$}
\end{algorithm}

\begin{corollary}
\label{corollary_1}
Assume that the optimal dual solution $\bm{y}^*$ of problem ({\ref{prob_primal}}) satisfies ${\left\| {{\bm{y}^*}} \right\|_2} \le R$. Then, for given tolerance $\varepsilon > 0$, the number of required iterations is $O\left({\left\| \bm{A} \right\|_2}\max \left( \bm{u} \right)\sqrt {\theta R/\varepsilon } \right)$.
\end{corollary}

\subsection{Backward pass in GLinSAT} \label{sec3.3}
Since all the operations involved in Algorithm {\ref{alg_apdagd}} are differentiable with respect to $\bm{c}$, a natural idea is to directly use the auto differential mechanism to calculate the derivatives in the backward pass. However, directly backward propagation may require ever growing memory to store computational graphs and may cost much time when the forward pass requires a lot of iteration steps. To save the memory usage and accelerate the derivative calculation, we also provide an alternative way based on the optimality condition to calculate the derivatives in GLinSAT.

First, by calculating the derivative of $-g\left(\bm{y}\right)$, we can obtain the optimality condition as follows:
\begin{equation}
	\label{kkt_eq}
	\bm{h}\left(\bm{y}\right) = \bm{A} \left(\bm{u} \circ \bm{\sigma} \left( { - {\theta}\bm{u} \circ \left(- \bm{c} - \bm{A}^T \bm{y}\right)} \right) \right) - \bm{b} = \bm{0}
\end{equation}
According to implicit differentiation and chain rule, differentiating equation ({\ref{kkt_eq}}), we can get:
\begin{equation}
	\frac{{\partial \bm{y}}}{{\partial \bm{c}}} = - {\left( {\frac{{\partial \bm{h}}}{{\partial \bm{y}}}} \right)^{ - 1}}{\frac{{\partial \bm{h}}}{{\partial \bm{c}}}}
\end{equation}
Furthermore, according to equation ({\ref{optimal_x}}), the derivative of loss function $l$ with respect to $\bm{c}$ can be calculated as:
\begin{equation}
	\frac{{\partial l}}{{\partial \bm{c}}} = \frac{{\partial l}}{{\partial \bm{x}}}\frac{{\partial \bm{x}}}{{\partial \bm{c}}} + \frac{{\partial l}}{{\partial \bm{x}}}\frac{{\partial \bm{x}}}{{\partial \bm{y}}}\frac{{\partial \bm{y}}}{{\partial \bm{c}}} + \frac{{\partial l}}{{\partial \bm{y}}}\frac{{\partial \bm{y}}}{{\partial \bm{c}}} = \frac{{\partial l}}{{\partial \bm{x}}}\frac{{\partial \bm{x}}}{{\partial \bm{c}}} - \left( {\frac{{\partial l}}{{\partial \bm{x}}}\frac{{\partial \bm{x}}}{{\partial \bm{y}}} + \frac{{\partial l}}{{\partial \bm{y}}}} \right){\left( {\frac{{\partial \bm{h}}}{{\partial \bm{y}}}} \right)^{ - 1}}\frac{{\partial \bm{h}}}{{\partial \bm{c}}}
\end{equation}
In the actual implementation of GLinSAT, we do not explicitly form these Jacobian matrices $\frac{{\partial \bm{x}}}{{\partial \bm{c}}}, \frac{{\partial \bm{x}}}{{\partial \bm{y}}}, \left(\frac{{\partial \bm{h}}}{{\partial \bm{c}}}\right)^{-1}\frac{{\partial \bm{h}}}{{\partial \bm{c}}}$. Instead, we directly form the matrix-vector products $\frac{{\partial l}}{{\partial \bm{x}}}\frac{{\partial \bm{x}}}{{\partial \bm{c}}}, \frac{{\partial l}}{{\partial \bm{x}}}\frac{{\partial \bm{x}}}{{\partial \bm{y}}}$. In addition, since the jacobian matrix $\frac{{\partial \bm{h}}}{{\partial \bm{y}}}$ is positive semi-definite (see derivations in Appendix {\ref{appendix.a4}}), we can use conjugate gradient method to calculate the inverse-matrix-vector products in GLinSAT. Therefore, only matrix-vector product operations are involved in the calculation of derivatives. Moreover, for the sake of completeness, in GLinSAT, we also implement derivatives with respect to $\bm{A}$, $\bm{b}$, $\bm{u}$ for future potential usage. The detailed derivation process of all derivatives is provided in Appendix {\ref{appendix.a4}}.

\section{Experimental Results}
In this section, experiments on constrained traveling salesman problems, partial graph matching with outliers, predictive portfolio allocation and power system unit commitment are used to demonstrate the advantages of GLinSAT through comparison with the state-of-the-art linear satisfiability layers LinSAT {\cite{wang2023linsatnet}}, CvxpyLayers {\cite{agrawal2019differentiable}} and OptNet {\cite{amos2017optnet}}. 
For OptNet, LinSAT and GLinSAT, the regularization coefficients of nonlinear terms are all set to $1 / \theta$. For CvxpyLayers, the projection problem to be solved is set to the same as ({\ref{prob_primal}}). 
The first three experiments originate from Ref. {\cite{wang2023linsatnet}}. The last experiment is the unit commitment problem in actual power systems. In the following sections, GLinSAT-(Dense/Sparse)-(Explicit/Implicit) means that GLinSAT is used with dense/sparse matrix and backpropagation is performed using automatic differential/implicit differential. LinSAT-(Dense/Sparse)-(100/500) means that LinSAT is used with dense/sparse matrix and maximum iteration number is set to 100/500. The reason we cannot set the maximum iteration number in LinSAT to $+\infty$, as we do in GLinSAT, is that LinSAT may iterate endlessly and get stuck in such a case. All the experiments are conducted on a computer with a 24-core Intel(R) Xeon(R) Platinum 8360H CPU and a NVIDIA Tesla A100 GPU through Pytorch 2.2. Our code is provided in {\url{https://github.com/HunterTracer/GLinSAT}}.

\subsection{Constrained traveling salesman problem} \label{sec4.1}
Using the traveling salesman problem (TSP) dataset in {\cite{wang2023linsatnet}}, here we test the performance of each satisfiability layer through experiments on TSP with starting and ending cities constraint and priority constraint respectively. The mathematical formulation of TSP with starting and ending cities constraint (TSP-StartEnd) and TSP with priority constraint (TSP-Priority) is provided in Appendix {\ref{appendix.a5}}.
%
The detailed experimental settings are provided in Appendix {\ref{appendix.a5}}. We report the average batch processing performance in Table {\ref{tab_performance_tsp}} where $\frac{1}{\theta}$ is set to 0.1. The results when $\frac{1}{\theta}$ is set to 10\textsuperscript{-2} are similar therefore we display the results in Table {\ref{tab_performance_tsp_2}} and Table {\ref{tab_result_tsp_2}} in Appendix {\ref{appendix.a5}}.

\begingroup
\renewcommand{\arraystretch}{1.0}
\begin{table}[h]
	\small
	\centering
	\caption{Average allocated GPU memory and solution time of different satisfiability layers during batch processing of projection and backpropagation when $\frac{1}{\theta}$ is set to 0.1 in TSP training phase}
	\begin{tabularx}{\textwidth}{M{0.234\textwidth}M{0.0426\textwidth}CM{0.0426\textwidth}CM{0.0426\textwidth}CM{0.0426\textwidth}C}
		\toprule
		& \multicolumn{4}{c}{TSP-StartEnd}       & \multicolumn{4}{c}{TSP-Priority} \\
		\cmidrule(lr){2-5} \cmidrule(lr){6-9}
		& \multicolumn{2}{c}{GPU Mem./MB} & \multicolumn{2}{c}{Time/s} & \multicolumn{2}{c}{GPU Mem./MB} & \multicolumn{2}{c}{Time/s} \\
		\cmidrule(lr){2-3} \cmidrule(lr){4-5} \cmidrule(lr){6-7} \cmidrule(lr){8-9}
		Layer & Proj. & Backprop.  & Proj. & Backprop.  & Proj. & Backprop.  & Proj. & Backprop.   \\
		\midrule
		CvxpyLayers   & \---\---     & \---\---    & 112.1     & 18.39     & \---\---     & \---\---     & 116.5     & 19.94 \\
		OptNet  & 14305     & 5005    & 18.92    & 0.929    & 14333    & 5005    & 20.26    & 1.136 \\
		LinSAT-Dense-100 & 14977    & 181.2    & \textbf{0.278}    & 0.417    & 15009    & 180.9    & \textbf{0.276}    & 0.418 \\
		LinSAT-Dense-500 & 74108    & 181.2    & 1.323    & 1.927    & 74272    & 180.9    & 1.317    & 1.929 \\
		GLinSAT-Dense-Explicit & 4380    & \textbf{4.868}    & 0.382    & 0.240    & 4898    & \textbf{4.880}    & 0.440    & 0.270 \\
		\cellcolor{gray!20}GLinSAT-Dense-Implicit & \cellcolor{gray!20}{\textbf{13.35}}    & \cellcolor{gray!20}{53.23}    & \cellcolor{gray!20}{0.306}    & \cellcolor{gray!20}{\textbf{0.143}}    & \cellcolor{gray!20}{\textbf{13.36}}    & \cellcolor{gray!20}{53.22}    & \cellcolor{gray!20}{0.349}    & \cellcolor{gray!20}{\textbf{0.146}} \\
		LinSAT-Sparse-100 & 7971    & 132.0    & 0.281    & 0.358    & 8026    & 132.6    & 0.286    & 0.368 \\
		LinSAT-Sparse-500 & 39020    & 133.5    & 1.356    & 1.614    & 39309    & 133    & 1.397    & 1.645 \\
		GLinSAT-Sparse-Explicit & 2787    & 5.426    & 0.603    & 0.326    & 3127    & 5.983    & 0.652    & 0.355 \\
		GLinSAT-Sparse-Implicit & 62.95    & 24.71    & 0.454    & 0.158    & 63.27    & 24.51    & 0.495    & 0.165 \\
		\bottomrule
	\end{tabularx}%
	\begin{tablenotes}
	\item[a] Note: The GPU memory used by CvxpyLayers is not counted since CvxpyLayers use the CPU parallel mechanism. Statistics of CvxpyLayers and OptNet are based on the first epoch since we cannot obtain a well-trained model in reasonable time.
	\end{tablenotes}
	\label{tab_performance_tsp}%
\end{table}%
\endgroup

From Table {\ref{tab_performance_tsp}}, it can be seen that GLinSAT-Dense-Implicit outperforms all the other methods with minimum total storage and shortest total computation time. We can also find that our proposed GLinSAT is memory-efficient. Even though we choose the GLinSAT-Dense-Explicit method which will cost the most memory among all versions of GLinSAT, the total memory usage is still less than that of LinSAT-Sparse-100. We also attempted to set the maximum number of iterations for LinSAT to $+\infty$, but at this point LinSAT will get stuck and we cannot obtain a reasonable result. Contrarily, for our proposed GLinSAT, the convergence is guaranteed, so setting the maximum iteration number to $+\infty$ will not affect the result. 

To obtain feasible tours, we exploit two kinds of post-processing methods in the validation stage {{\cite{wang2023linsatnet}}}. The first one is rounding and the second one is beam search where the width of the beam is set to 2048. Table {\ref{tab_result_tsp}} shows the average tour length and feasibility ratio of each method. Since CvxpyLayers and OptNet are hundreds of times slower than LinSAT and GLinSAT, we are unable to obtain trained models within reasonable time and thus the results are not included.

\begingroup
\renewcommand{\arraystretch}{1.0}
\begin{table}[htbp]
	\small
	\centering
	\caption{Mean tour length and feasibility ratio obtained from using different $\frac{1}{\theta}$ and post-processing methods in TSP validation stage}
	\begin{tabularx}{\textwidth}{M{0.234\textwidth}M{0.0625\textwidth}M{0.0625\textwidth}M{0.0625\textwidth}M{0.0625\textwidth}M{0.0625\textwidth}M{0.0625\textwidth}M{0.0625\textwidth}M{0.0625\textwidth}}
		\toprule
		& \multicolumn{4}{c}{TSP-StartEnd}       & \multicolumn{4}{c}{TSP-Priority} \\
		\cmidrule(lr){2-5} \cmidrule(lr){6-9}
		& \multicolumn{2}{c}{\makecell{Rounding \\with $\frac{1}{\theta}$ = 10\textsuperscript{-2} }} & \multicolumn{2}{c}{\makecell{Beamsearch \\with $\frac{1}{\theta}$ = 10\textsuperscript{-1} }} & \multicolumn{2}{c}{\makecell{Rounding \\with $\frac{1}{\theta}$ = 10\textsuperscript{-2} }} & \multicolumn{2}{c}{\makecell{Beamsearch \\with $\frac{1}{\theta}$ = 10\textsuperscript{-1} }} \\
		\cmidrule(lr){2-3} \cmidrule(lr){4-5} \cmidrule(lr){6-7} \cmidrule(lr){8-9}
		Layer  & Mean Length & Feas. Ratio & Mean Length & Feas. Ratio & Mean Length & Feas. Ratio & Mean Length & Feas. Ratio \\
		\midrule
		LinSAT-Dense-100 & 4.007    & 15.3\%    & 3.843    & \textbf{100\%}    & 4.114    & 41.6\%    & 3.952    & \textbf{100\%} \\
		LinSAT-Dense-500 & 3.926    & 93.6\%    & 3.823    & \textbf{100\%}    & 4.098    & 91.5\%    & 3.947    & \textbf{100\%} \\
		GLinSAT-Dense-Explicit & 3.939    & 94.2\%    & 3.817    & \textbf{100\%}   & 4.079    & \textbf{93.5\%}    & 3.934    & \textbf{100\%} \\
		\cellcolor{gray!20}GLinSAT-Dense-Implicit & \cellcolor{gray!20}\textbf{3.922}    & \cellcolor{gray!20}94.2\%    & \cellcolor{gray!20}\textbf{3.811}    & \cellcolor{gray!20}\textbf{100\%}    & \cellcolor{gray!20}\textbf{4.068}    & \cellcolor{gray!20}93.4\%    & \cellcolor{gray!20}\textbf{3.927}    & \cellcolor{gray!20}\textbf{100\%} \\
		LinSAT-Sparse-100 & \---\---    & \---\---    & 3.843    & \textbf{100\%}    & \---\---    & \---\---    & 4.567    & \textbf{100\%} \\
		LinSAT-Sparse-500 & \---\---    & \---\---    & 3.818    & \textbf{100\%}    & \---\---    & \---\---    & 4.400    & \textbf{100\%} \\
		GLinSAT-Sparse-Explicit & 3.939    & 94.2\%    & 3.817    & \textbf{100\%}    & 4.078    & 92.6\%    & 3.935    & \textbf{100\%} \\
		GLinSAT-Sparse-Implicit & 3.929    & \textbf{94.6\%}    & 3.818    & \textbf{100\%}    & 4.073    & 93.4\%    & 3.933    & \textbf{100\%} \\
		\bottomrule
	\end{tabularx}%
	\begin{tablenotes}
		\item[a] Note: The output of LinSAT-Sparse when $1/\theta$ = 10\textsuperscript{-2} is not a real number so that the results are not shown.
	\end{tablenotes}
	\label{tab_result_tsp}%
\end{table}
\endgroup

From Table {\ref{tab_result_tsp}}, we can see that GLinSAT-Dense-Implicit results in the shortest mean tour length when beamsearch is used. It is also noteworthy that LinSAT will produce poor solution when we apply rounding to the results and the max iteration number is set to 100. Although setting the maximum number of iterations to 500 can improve LinSAT's performance, LinSAT's performance is still not as good as GLinSAT. Considering that the total computation time of LinSAT is more than five times that of GLinSAT at this time, it makes LinSAT less competitive.
%
%
\subsection{Partial graph matching with outliers} \label{sec4.2}
The detailed mathematical formulation of partial graph matching with outliers is provided in {\ref{appendix.a6}}. We carry out experiments on Pascal VOC Keypoint dataset {\cite{everingham2010pascal}} with Berkeley annotations {\cite{bourdev2009poselets}} under the unfiltered setting {\cite{wang2023linsatnet, rolinek2020deep}}.

Considering there are graphs with different sizes in one batch, we stack constraints as block diagonal matrices and forward them to LinSAT and GLinSAT. However, CvxpyLayers and OptNet currently cannot handle large block diagonal matrices. Disciplined parameterized programming compilation in CvxpyLayers and matrix factorization of large matrices in OptNet will cost a significant amount of time. Therefore, we can only use a for-loop to handle a batch with different sizes separately. The average GPU memory usage and solution time across different satisfiability layers is provided in Table {\ref{tab_performance_match}} of Appendix {\ref{appendix.a6}}. In the validation stage, we use Hungarian algorithm and greedy strategy for obtaining feasible integer solutions {\cite{wang2023linsatnet}}. We regard the cost of matching a pair of nodes as the outputs of satisfiability layers, then use Hungarian algorithm to obtain a maximum matching. Finally, we use greedy strategy to preserve pairs with 
top-$p$ matching scores for constraint satisfaction. The matching F1 scores between graph pairs across various satisfiability layers are shown in Table {\ref{tab_result_match}}. The result of LinSAT-Dense-500 is not given due to out-of-memory (OOM) issues. According to Table {\ref{tab_result_match}}, we can find GLinSAT yields the highest F1 scores across all satisfiability layers.

\begingroup
\renewcommand{\arraystretch}{1.0}
\begin{table}[htbp]
	\small
	\centering
	\caption{Mean F1 scores across different satisfiability layers in partial graph matching problem}
	\begin{tabularx}{\textwidth}{cM{0.09\textwidth}M{0.09\textwidth}cM{0.09\textwidth}M{0.09\textwidth}}
		\toprule
		& $\frac{1}{\theta}$ = 10\textsuperscript{-1} & $\frac{1}{\theta}$ = 10\textsuperscript{-2} &       & $\frac{1}{\theta}$ = 10\textsuperscript{-1} & $\frac{1}{\theta}$ = 10\textsuperscript{-2} \\
		\cmidrule(lr){2-2} \cmidrule(lr){3-3} \cmidrule(lr){5-5} \cmidrule(lr){6-6}
		Layer & Mean F1 & Mean F1 & Layer & Mean F1 & Mean F1 \\
		\midrule
		CvxpyLayers   & 0.616      &  0.605     & OptNet   &  0.619     & 0.613  \\
		LinSAT-Dense-100  & 0.619      & 0.614  & LinSAT-Dense-500  & $\times$      & $\times$ \\
		\cellcolor{gray!20}GLinSAT-Dense-Explicit & \cellcolor{gray!20}\textbf{0.620}      & \cellcolor{gray!20}\textbf{0.620}      & \cellcolor{gray!20}GLinSAT-Dense-Implicit & \cellcolor{gray!20}0.619      & \cellcolor{gray!20}\textbf{0.620} \\
		LinSAT-Sparse-100 & \textbf{0.620}  & 0.618 & LinSAT-Sparse-500 & 0.619 & 0.611 \\
		\cellcolor{gray!20}GLinSAT-Sparse-Explicit & \cellcolor{gray!20}0.619      & \cellcolor{gray!20}\textbf{0.620}      & \cellcolor{gray!20}GLinSAT-Sparse-Implicit & \cellcolor{gray!20}\textbf{0.620}      & \cellcolor{gray!20}\textbf{0.620} \\
		\bottomrule
	\end{tabularx}%
	\label{tab_result_match}%
\end{table}%
\endgroup

\subsection{Predictive portfolio allocation} \label{sec4.3}
In this section, we use the predictive portfolio allocation dataset in {\cite{wang2023linsatnet}}. Denote $x_i \in \left[0, 1\right]$ as the predicted portfolio decision variable of asset $i$, $\cal S$ as the preferred portfolio asset. Our portfolio allocation needs to maximize the Sharpe ratio {\cite{sharpe1998sharpe}} while ensuring decision variables satisfy constraints $\sum_{i = 1}^n {{x_i}}  = 1, \sum_{i \in \cal S}^{} {{x_i}}  \ge q$ where $q$ is a pre-defined positive constant.
The details of experiments are provided in Appendix {\ref{appendix.a7}}. The average memory usage and solution time is shown in Table {\ref{tab_performance_port}} of Appendix {\ref{appendix.a7}}. According to Table {\ref{tab_performance_port}}, we can find that our proposed GLinSAT is the fastest layer among all layers. In Table {\ref{tab_result_port}}, we show the mean Sharpe ratio obtained from different satisfiability layers. Our proposed method always yields a high Sharpe ratio whether $\theta$ takes 10\textsuperscript{-1} or 10\textsuperscript{-2}.
\begingroup
\renewcommand{\arraystretch}{1.0}
\begin{table}[htbp]
	\small
	\centering
	\caption{Mean Sharpe ratio obtained from different satisfiability layers in portfolio allocation problem}
	\begin{tabularx}{\textwidth}{cM{0.09\textwidth}M{0.09\textwidth}cM{0.09\textwidth}M{0.09\textwidth}}
		\toprule
		& $\frac{1}{\theta}$ = 10\textsuperscript{-1} & $\frac{1}{\theta}$ = 10\textsuperscript{-2} &       & $\frac{1}{\theta}$ = 10\textsuperscript{-1} & $\frac{1}{\theta}$ = 10\textsuperscript{-2} \\
		\cmidrule(lr){2-2} \cmidrule(lr){3-3} \cmidrule(lr){5-5} \cmidrule(lr){6-6}
		Layer & S. Ratio & S. Ratio & Layer & S. Ratio & S. Ratio \\
		\midrule
		CvxpyLayers   & 2.535      &  \textbf{2.600}     & OptNet   &  \textbf{2.553}     & 2.381  \\
		LinSAT-Dense-100  & 2.245      & 2.409  & LinSAT-Dense-500  & 2.245      & 2.409 \\
		\cellcolor{gray!20}GLinSAT-Dense-Explicit & \cellcolor{gray!20}2.535      & \cellcolor{gray!20}\textbf{2.608}      & \cellcolor{gray!20}GLinSAT-Dense-Implicit & \cellcolor{gray!20}2.535      & \cellcolor{gray!20}\textbf{2.608} \\
		LinSAT-Sparse-100 & 2.245  & 2.409 & LinSAT-Sparse-500 & 2.245 & 2.409 \\
		\cellcolor{gray!20}GLinSAT-Sparse-Explicit & \cellcolor{gray!20}2.535      & \cellcolor{gray!20}\textbf{2.608}      & \cellcolor{gray!20}GLinSAT-Sparse-Implicit & \cellcolor{gray!20}2.535      & \cellcolor{gray!20}\textbf{2.608} \\
		\bottomrule
	\end{tabularx}%
	\label{tab_result_port}%
\end{table}%
\endgroup

\subsection{Power system unit commitment} \label{sec4.4}
In this section, we carry out experiments about the unit commitment problem on a real provincial power system. In the unit commitment problem, there are hard constraints and soft constraints. Generally, constraints directly related to generators are regarded as hard constraints, e.g. the generator logical constraints, generator minimum up-time and down-time constraints. Constraints related to the section power and load balance are usually regarded as soft constraints where violation with large penalty coefficient is introduced in the objective {\cite{wu2021novel}}. The unit commitment problem can be formulated into a mixed-integer linear programming (MILP) problem, which is detailed in Appendix {\ref{appendix.a8}}. Based on one-year power system load data, we first use Gurobi {\cite{refGurobi}} to solve the MILP within a 0.1\% optimality gap. 

After obtaining the integer solution of the unit commitment problem, we then use supervised learning to train neural networks with satisfiability layers so that they can predict the optimal state of a unit while satisfying logical constraints, minimum up-time and down-time constraints. As pointed by {\cite{rajan2005minimum}}, when we consider logical constraints and minimum up-time and down-time constraints, these constraints formulate a convex hull so that the extreme points of the corresponding feasible region are binary. As a result, we could expect that the outputs of satisfiability layers tend to be binary when $1/\theta \to 0$, thereby making all constraints, including integer constraints, more likely to be satisfied after rounding operations. Once we obtain the predicted integer commitment status of the generators, we can fix the integer variables in the unit commitment problem and solve the corresponding linear programming problem, thereby providing a good initial point for the original mixed-integer programming problem.

Since negative coefficients occur in constraints, LinSAT cannot be used. In Table {\ref{tab_performance_uc}} of Appendix {\ref{appendix.a8}}, we compare the performance of batch processing with different layers. When we stack constraints into block diagonal form to exploit parallelism, there are about 1000000 rows and 2000000 columns in the matrix. GLinSAT-Sparse-Implicit is the only way that will not report out-of-memory issues when we use GLinSAT. Both CvxpyLayers and OptNet cannot directly handle such a giant matrix within reasonable time thus we can only use a sequential way instead.

We train neural networks with $\frac{1}{\theta} = 0.1$. Table {\ref{tab_result_uc}} shows the feasibility ratio and average gap on feasible solutions obtained from fixing unit state variables to rounded outputs of neural networks and then solving the continuous unit commitment problem in validation stage. Since CvxpyLayers and OptNet are significantly slower than GLinSAT, we are unable to obtain trained models within reasonable time and the results are not included in Table {\ref{tab_result_uc}}. Table {\ref{tab_result_uc}} shows that if we use sigmoid function to replace the satisfiability layer in training and validation, we cannot obtain any feasible solution. As $\frac{1}{\theta} \to 0$, the feasibility ratio increases. When using GLinSAT with $\frac{1}{\theta} \le 0.0005$, the feasibility ratio reaches 100\%. In addition, when we set $\frac{1}{\theta}$ to exactly zero and solve the resulted projection problem in the form of linear programming (LP) via Gurobi, 100\% feasible solutions are also found.
\begingroup
\renewcommand{\arraystretch}{1.0}
\begin{table}[htbp]
	\small
	\centering
	\caption{Feasibility ratio and average gap obtained from using different $1 / \theta$ in validation}
	\begin{tabularx}{\textwidth}{M{0.234\textwidth}M{0.234\textwidth}M{0.08\textwidth}CC}
		\toprule
		Training final layer & Validation final layer & $1 / \theta$ & Feasibility Ratio & Average Gap \\
		\midrule
		Sigmoid   & Sigmoid     &  \---\---     & 0\%   & \---\---  \\
		GLinSAT-Sparse-Implicit  & GLinSAT-Sparse-Implicit      & 0.01  & 86.23\% & 0.1119\%  \\
		GLinSAT-Sparse-Implicit  & GLinSAT-Sparse-Implicit      & 0.005  & 95.41\% & 0.1381\% \\
		GLinSAT-Sparse-Implicit  & GLinSAT-Sparse-Implicit      & 0.001  & 98.17\% & \textbf{0.1109\%} \\
		\cellcolor{gray!20}GLinSAT-Sparse-Implicit  & \cellcolor{gray!20}GLinSAT-Sparse-Implicit      & \cellcolor{gray!20}0.0005  & \cellcolor{gray!20}\textbf{100\%} & \cellcolor{gray!20}0.1114\% \\
		\cellcolor{gray!20}GLinSAT-Sparse-Implicit  & \cellcolor{gray!20}GLinSAT-Sparse-Implicit      & \cellcolor{gray!20}0.0001  & \cellcolor{gray!20}\textbf{100\%} & \cellcolor{gray!20}0.1114\% \\
		\cellcolor{gray!20}GLinSAT-Sparse-Implicit  & \cellcolor{gray!20}Gurobi-LP      & \cellcolor{gray!20}0  & \cellcolor{gray!20}\textbf{100\%} & \cellcolor{gray!20}0.1114\% \\
		\bottomrule
	\end{tabularx}%
	\label{tab_result_uc}%
	\vspace{-1em}
\end{table}%
\endgroup

\section{Conclusion}
In this paper, we reformulate the neural network output projection problem into a convex optimization problem with Lipschitz continuous gradient. We then propose GLinSAT, a general linear satisfiability layer to impose linear constraints on neural network outputs where all the operations are differentiable and matrix-factorization-free. GLinSAT can fully leverage the parallel computing capabilities of the GPU. We showcase four applications of GLinSAT and the advantages of our proposed framework over existing satisfiability layers are illustrated.

\begin{ack}
This work was supported in part by the National Natural Science
Foundation of China under Grant U22B2097, 52321004 and in part by Alibaba Innovative Research Program. We would like to express our sincerest gratitude to the anonymous reviewers for their insightful feedback on our work. We are also immensely thankful to Wotao Yin for his invaluable support throughout the research process.
\end{ack}

\bibliography{refs}

%
%
%
%
%
%
%
%
%

\newpage
\appendix

\section{Appendix}
\numberwithin{equation}{section}
\setcounter{equation}{0}
\numberwithin{table}{section}
\setcounter{table}{0}
\subsection{Related Work}
\textbf{Constraints handling paradigm in neural networks for decision making}. Some simple constraints can be directly encoded by neural network activation functions, e.g., ReLU function for non-negative constraints, sigmoid function for bounded constraints, softmax function for sum-to-one constraints. However, it is difficult for neural network outputs to satisfy complicated constraints by only using these activation functions. For a few problems with special structures, the constraints can be directly handled by the well-designed action space of sequential decisions in reinforcement learning {\cite{bello2016neural, khalil2017learning}}. However, as the complexity of constraints increases, it will be hard to design a suitable action space. Considering the difficulty for agents to find feasible solutions through random exploration, the resulting sparse rewards may also lead to slow convergence. Another common way to deal with constraint violation is to penalize such violation in the training stage. Ref. {\cite{karalias2020erdos}} incorporates the constraint violation in the loss function of supervised learning. Ref. {\cite{bohez2019value, bhatia2019resource}} augment the reward function in reinforcement learning with the sum of the constraint violation penalty weighted by the Lagrange multipliers. However, it cannot be guaranteed that the constraints can be always satisfied by directly penalizing the constraint violation, see Ref. {\cite{wang2022towards}}. Ref. {\cite{zhao2023ensuring, tabas2022computationally, tordesillas2023rayen}} are better suited for inequality constraint satisfaction. These methods may encounter difficulties in satisfying equality constraints, either in terms of efficiency or expressiveness. Another way to handle constraints is to incorporate optimization solvers into neural network layers, which is detailed in the next paragraph.

\textbf{Optimizers as neural network layers for constraint satisfaction}. To integrate optimizers into neural networks, it is necessary to calculate derivatives with respect to the parameters and perform batch processing efficiently. Ref. {\cite{poganvcic2019differentiation, berthet2020learning, paulus2021comboptnet}} exploit black-box solvers to impose combinatorial constraints on decision variables. However, only approximated gradients can be obtained through perturbations on the problem. Ref {\cite{wang2019satnet}} relax the original combinatorial constraint into the positive semi-definite matrix constraint with unit diagonals and integrates GPU-based MAXSAT solver into neural network layers. In Ref. {\cite{santa2017deeppermnet, wang2019learning, cuturi2019differentiable}}, Sinkhorn algorithm {\cite{sinkhorn1967concerning, cuturi2013sinkhorn, chakrabarty2021better}} is used to make neural network outputs satisfy the double stochastic matrix constraint, which is a linear relaxation of permutation, matching and sorting constraints. Since Sinkhorn algorithm only involves iterative normalization of rows and columns in matrices, it is straightforward to parallel the operations on GPUs and calculate the derivatives based on automatic differential mechanism. Ref. {\cite{amos2017optnet}} presents OptNet, a neural network layer that integrates a GPU-based batched quadratic programming solver, qpth. The forward pass exploits a primal-dual interior point method to find the solution and the derivatives are calculated based on the matrix factorization obtained from the forward propagation. Ref. {\cite{agrawal2019differentiable}} further presents CvxpyLayers to incorporate convex programming into neural network layers. 
Although CvxpyLayers can represent more optimization problems, it relies on CPUs for parallelism which may lead to efficiency issues if a large batch of optimization problems needs solving. Ref. {\cite{wang2023linsatnet}} designs LinSAT, a differentiable layer to encode the positive linear constraints based on a multi-set Sinkhorn algorithm. Although such an algorithm is easy to be parallel on GPUs, it can only imposing positive linear constraints on neural network outputs.

\subsection{Broader Impacts}
\label{broader_impacts}
This paper is aimed at making the outputs of neural networks satisfy bounded and general linear constraints. The proposed framework GLinSAT can be used for end-to-end neural network training while ensuring the feasibility of neural network outputs, providing a promising approach for applying neural networks to decision-making problems. There is no foreseeable negative societal consequence that is a direct result of the proposed method.

\subsection{Limitations}
\label{limitations}
In this paper, we manage to impose bounded and linear constraints on neural network outputs. Considering that real-life decision variables often have finite upper and lower bounds simultaneously, our proposed method can actually be applied to a lot of decision-making problems. However, for variables with one-sided boundary or no explicit boundary, our method cannot be directly used. A possible workaround is to manually calculate the implicit bounds of these variables through domain propagation, see {\cite{savelsbergh1994preprocessing, fugenschuh2005computational, achterberg2007constraint, achterberg2020presolve, gleixner2023papilo}}. In addition, it should be noted that currently our algorithm can only deal with linear constraints. In the future, we need to conduct further research on neural network layers that can efficiently handle cone constraints on GPUs.

\subsection{Reformulation of general linear constraints with bounded variables into standard form} \label{appendix.a1}
Denote $\bm{c}' \in \mathbb{R}^{n'}$ as the output of an neural network. In this section, we consider projecting the output of the neural network $\bm{x}' \in \mathbb{R}^{n'}$ into variables $\bm{x}' \in \mathbb{R}^{n'}$ that satisfy the following linear constraints and bounded constraints:
\begin{subequations}
	\begin{gather}
			\bm{A}_1' \bm{x}' \leq \bm{b}_1' \\
			\bm{A}_2' \bm{x}' \geq \bm{b}_2' \\
			\bm{A}_3' \bm{x}' = \bm{b}_3' \\
			\bm{l}' \leq \bm{x}' \leq \bm{u}'
		\end{gather}
\end{subequations}
where $\bm{A}_1' \in \mathbb{R}^{m_1' \times {n'}}$, $\bm{b}_1' \in \mathbb{R}^{m_1'}$, $\bm{A}_2' \in \mathbb{R}^{m_2' \times {n'}}$, $\bm{b}_2' \in \mathbb{R}^{m_2'}$, $\bm{A}_3' \in \mathbb{R}^{m_3' \times {n'}}$, $\bm{b}_3' \in \mathbb{R}^{m_3'}$, $\bm{l} \in \mathbb{R}^{n'}$, $\bm{u}' \in \mathbb{R}^{n'}$.

Obviously, we can convert all inequality constraints into equality constraints by introducing bounded slack variables as follows:
\begin{subequations}
	\begin{gather}
			\bm{A}_1' \bm{x}' + \bm{s}_1' = \bm{b}_1 \\
			\bm{A}_2' \bm{x}' - \bm{s}_2' = \bm{b}_2 \\
			\bm{A}_3' \bm{x}' = \bm{b}_3' \\
			\bm{l}' \leq \bm{x}' \leq \bm{u}' \\
			\bm{0} \leq \bm{s}_1' \leq \overline{\bm{s}_1'} \\
			\bm{0} \leq \bm{s}_2' \leq \overline{\bm{s}_2'}
		\end{gather}
\end{subequations}
where $\overline{\bm{s}_1'} = \bm{b}_1 - {\bm{A}_{1}^{+'}} \bm{l}' - {\bm{A}_{1}^{-'}} \bm{u}'$, $\overline{\bm{s}_2'} = {\bm{A}_{2}^{+'}} \bm{u}' + {\bm{A}_{2}^{-'}} \bm{l}' - \bm{b}_2$, $\bm{A}_{1+}', \bm{A}_{2+}'$ and $\bm{A}_{1-}', \bm{A}_{2-}'$ are the positive and negative parts of matrix $\bm{A}_{1}', \bm{A}_{2}'$ respectively.

We use the following notation:\[ \bm{x} = \left[ {\begin{array}{*{20}{c}}
		{\bm{x}' - \bm{l}'}\\
		{{\bm{s}_1'}}\\
		{{\bm{s}_2'}}
\end{array}} \right], \bm{A} = \left[ {\begin{array}{*{20}{c}}
		{{\bm{A}_1'}}&\bm{I}&{}\\
		{{\bm{A}_2'}}&{}&{ - \bm{I}}\\
		{{\bm{A}_3'}}&{}&{}
\end{array}} \right], \bm{b} = \left[ {\begin{array}{*{20}{c}}
		{{\bm{b}_1'} - {\bm{A}_1'}\bm{l}'}\\
		{{\bm{b}_2'} - {\bm{A}_2'}\bm{l}'}\\
		{{\bm{b}_3'}}
\end{array}} \right], \bm{c} = \left[ {\begin{array}{*{20}{c}}
		\bm{c}'\\
		\bm{0}\\
		\bm{0}
\end{array}} \right], \bm{u} = \left[ {\begin{array}{*{20}{c}}
		{\bm{u}' - \bm{l}'}\\
		{\overline{\bm{s}_1'}}\\
		{\overline{\bm{s}_2'}}
\end{array}} \right]\]
Then, the original problem is transformed into a problem of projecting $\bm{c} \in \mathbb{R}^{n' + m_1' + m_2'}$ onto variables $\bm{x} \in \mathbb{R}^{n' + m_1' + m_2'}$ that satisfy the following linear constraints and bounded constraints:
\begin{subequations}
	\begin{gather}
			\bm{A} \bm{x} = \bm{b} \\
			\bm{0} \leq \bm{x} \leq \bm{u}
		\end{gather}
\end{subequations}

\subsection{Property of the primal and dual objective function} \label{appendix.a2}
We first show that the primal objective function $f\left(\bm{x}\right)$ is strongly convex, where $f\left(\bm{x}\right)$ is:
\begin{equation}
	f\left(\bm{x}\right) = - \bm{c}^T \bm{x} + \frac{1}{\theta }\sum\limits_{j = 1}^n {\left( {\frac{{{x_j}}}{{{u_j}}}\log \frac{{{x_j}}}{{{u_j}}} + \left( {1 - \frac{{{x_j}}}{{{u_j}}}} \right)\log \left( {1 - \frac{{{x_j}}}{{{u_j}}}} \right)} \right)}
\end{equation}

The second order derivative of $f\left(\bm{x}\right)$ can be expressed as follows:
\begin{equation}
	\nabla^2 f\left(\bm{x}\right) = \frac{1}{\theta}\bm{{\rm{diag}}}\left(\frac{\bm{1}}{\bm{x} \circ \left(\bm{u} - \bm{x}\right)}\right)
\end{equation}
where $\bm{{\rm{diag}}} \left(\cdot\right)$ maps a vector to its corresponding diagonal matrix, $\circ$ represents the element-wise product, $\frac{\bm{1}}{\bm{z}}$ represents the element-wise reciprocal of vector $\bm{z}$.

Since we have $\bm{0} \leq \bm{x} \leq \bm{u}$, we have $\bm{x} \circ \left(\bm{u} - \bm{x}\right) \leq \frac{\bm{u}}{2} \circ \frac{\bm{u}}{\bm{2}} = \frac{1}{4} \bm{u} \circ \bm{u}$. As a result, we have $\nabla^2 f\left(\bm{x}\right) \succeq \frac{4}{\theta \max\left({\bm{u}}\right)^2} \bm{I}$, which means $f$ is a $\frac{4}{\theta \max\left({\bm{u}}\right)^2}$-strongly convex function.

We next show that the gradient of the opposite dual objective function $\nabla \left(- g\left(\bm{y}\right)\right)$ is Lipschitz continuous, where $- g\left(\bm{y}\right)$ is:
\begin{equation}
	-g\left(\bm{y}\right) = - \frac{1}{{{\theta }}} \bm{1}^T {\bm{\log} {\bm{\sigma} \left( {{\theta }\bm{u} \circ \left(- \bm{c} - \bm{A}^T\bm{y}\right)} \right)}} - {\bm{b}^T}\bm{y}
\end{equation}

The first order derivative of $-g\left(\bm{y}\right)$ can be expressed as follows:
\begin{equation}
	\nabla \left(- g\left(\bm{y}\right)\right) = \bm{A} \bm{x}\left(\bm{y}\right) - \bm{b}
\end{equation}
where $\bm{x}\left(\bm{y}\right) = \bm{u} \circ \bm{\sigma} \left( { - {\theta}\bm{u} \circ \left(- \bm{c} - \bm{A}^T \bm{y}\right)} \right)$.

Using the first order derivative of $-g\left(\bm{y}\right)$, we have:
\begin{equation}
	\label{proof_lip_1}
	\left\| \nabla \left(-g\left(\bm{y}_1\right)\right) - \nabla \left(-g\left(\bm{y}_2\right)\right) \right\|_2 = \left\| \bm{A} \left(\bm{x}\left(\bm{y}_1\right) - \bm{x}\left(\bm{y}_2\right)\right) \right\|_2 \leq \left\| \bm{A} \right\|_2 \left\| \bm{x}\left(\bm{y}_1\right) - \bm{x}\left(\bm{y}_2\right) \right\|_2
\end{equation}

According to equation the strong convexity of $f\left(\bm{x}\right)$, we have:
\begin{equation}
	\begin{aligned}
		&\frac{4}{\theta \max\left({\bm{u}}\right)^2} {\left\| \bm{x}\left(\bm{y}_1\right) - \bm{x}\left(\bm{y}_2\right) \right\|_2^2} \leq \left(\nabla f\left(\bm{x}\left(\bm{y}_1\right)\right) - \nabla f\left(\bm{x}\left(\bm{y}_2\right)\right)\right)^T \left(\bm{x}\left(\bm{y}_1\right) - \bm{x}\left(\bm{y}_2\right)\right) \\
		& = \left((-\bm{c} + \frac{\bm{1}}{{\theta \bm{u}}}\circ \bm{\log} \frac{{\bm{x}\left( \bm{y}_1 \right)}}{{\bm{u} - \bm{x}\left( \bm{y}_1 \right)}}) - (-\bm{c} + \frac{\bm{1}}{{\theta \bm{u}}}\circ \bm{\log} \frac{{\bm{x}\left( \bm{y}_2 \right)}}{{\bm{u} - \bm{x}\left( \bm{y}_2 \right)}})\right)^T\left(\bm{x}\left(\bm{y}_1\right) - \bm{x}\left(\bm{y}_2\right)\right) \\
		& = \left(\bm{A}\left(\bm{y}_1 - \bm{y}_2\right)\right)^T \left(\bm{x}\left(\bm{y}_1\right) - \bm{x}\left(\bm{y}_2\right)\right) \leq \left\| \bm{A} \right\|_2 \left\| \bm{y}_1 - \bm{y}_2 \right\|_2 \left\| \bm{x}\left(\bm{y}_1\right) - \bm{x}\left(\bm{y}_2\right) \right\|_2
	\end{aligned}
\end{equation}
which implies:
\begin{equation}
	\label{proof_lip_2}
	\left\| \bm{x}\left(\bm{y}_1\right) - \bm{x}\left(\bm{y}_2\right) \right\|_2 \leq \frac{\theta \max\left(\bm{u}\right)^2 \left\| \bm{A} \right\|_2}{4} \left\| \bm{y}_1 - \bm{y}_2 \right\|_2
\end{equation}

Combining equation ({\ref{proof_lip_1}}) and ({\ref{proof_lip_2}}), we have:
\begin{equation}
	\left\| \nabla \left(-g\left(\bm{y}_1\right)\right) - \nabla \left(-g\left(\bm{y}_2\right)\right) \right\|_2 \leq \frac{\theta \max\left(\bm{u}\right)^2 \left\| \bm{A} \right\|_2^2}{4} \left\| \bm{y}_1 - \bm{y}_2 \right\|_2
\end{equation}
which means that $\nabla\left(-g\left(\bm{y}\right)\right)$ is $\frac{\theta \max\left(\bm{u}\right)^2 \left\| \bm{A} \right\|_2^2}{4}$-Lipschitz continuous.

\subsection{Time Complexity of Algorithm {\ref{alg_apdagd}}} \label{appendix.a3}
According to Theorem 2 in the supplementary material of {\cite{dvurechensky2018computational}},
the $l_2$-norm of $\bm{A}\bm{x}^{\left(k\right)} - \bm{b}$  at the $k$-th iteration is bounded by $16LR/k^2$, where $L$ is the Lipschitz constant of the gradient of the dual objective function, $R$ is the upper bound of $l_2$-norm of the optimal dual variables. Since we have shown that $\nabla\left(-g\left(\bm{y}\right)\right)$ is $\frac{\theta \max\left(\bm{u}\right)^2 \left\| \bm{A} \right\|_2^2}{4}$-Lipschitz in Appendix {\ref{appendix.a2}}, we can see that:
\begin{equation}
	\left\| \bm{A}\bm{x}^{\left(k\right)} - \bm{b} \right\|_2 \leq \frac{4 \theta \max\left(\bm{u}\right)^2 \left\| \bm{A} \right\|_2^2 R}{k^2}
\end{equation}

Let $\left\| \bm{A}\bm{x}^{\left(k\right)} - \bm{b} \right\|_2 = \varepsilon$, We can obtain the upper bound of outer cycle iteration number as follows:
\begin{equation}
	k \leq 2 \max\left(\bm{u}\right) \left\| \bm{A} \right\|_2 \sqrt{ \frac{\theta R} {\varepsilon}}
\end{equation}

According to Theorem 1 in the supplementary material of {\cite{dvurechensky2018computational}}, the number of all the iterations after an iteration $k$ is $O \left(k\right)$. Therefore, the time complexity of Algorithm {\ref{alg_apdagd}} is $O \left( \max\left(\bm{u}\right) \left\| \bm{A} \right\|_2 \sqrt{ {\theta R} / {\varepsilon}} \right)$. As a result, for optimization problems that with similar upper bounds of dual variables, the required iteration number is approximately proportional to $\sqrt{\theta / \varepsilon}$. Such time complexity is better than the complexity of sinkhorn-based algorithms, in which the iteration number is approximately proportional to $\theta / \varepsilon^2$, see {\cite{chakrabarty2021better, wang2023linsatnet}}.

\subsection{Derivative Calculation in the backward pass of GLinSAT} \label{appendix.a4}
The optimality condition can be expressed as follows:
\begin{equation}
	\label{kkt_eq_appendix}
	\bm{h}\left(\bm{y}\right) = \bm{A} \left(\bm{u} \circ \bm{\sigma} \left( { - {\theta}\bm{u} \circ \left(- \bm{c} - \bm{A}^T \bm{y}\right)} \right) \right) - \bm{b} = \bm{0}
\end{equation}
Let $\bm{v} = \bm{A}\; {\rm{or}}\; \bm{b}\; {\rm{or}}\; \bm{c}\; {\rm{or}}\; \bm{u}$, according to chain rule, we have:
\begin{equation}
	\label{derivative_chain}
	\frac{{\partial l}}{{\partial \bm{v}}} = \frac{{\partial l}}{{\partial \bm{x}}}\frac{{\partial \bm{x}}}{{\partial \bm{v}}} + \frac{{\partial l}}{{\partial \bm{x}}}\frac{{\partial \bm{x}}}{{\partial \bm{y}}}\frac{{\partial \bm{y}}}{{\partial \bm{v}}} + \frac{{\partial l}}{{\partial \bm{y}}}\frac{{\partial \bm{y}}}{{\partial \bm{v}}} = \frac{{\partial l}}{{\partial \bm{x}}}\frac{{\partial \bm{x}}}{{\partial \bm{v}}} - \left( {\frac{{\partial l}}{{\partial \bm{x}}}\frac{{\partial \bm{x}}}{{\partial \bm{y}}} + \frac{{\partial l}}{{\partial \bm{y}}}} \right){\left( {\frac{{\partial \bm{h}}}{{\partial \bm{y}}}} \right)^{ - 1}}\frac{{\partial \bm{h}}}{{\partial \bm{v}}}
\end{equation}
where $\bm{x}\left(\bm{y}\right) = \bm{u} \circ \bm{\sigma} \left( { - {\theta}\bm{u} \circ \left(- \bm{c} - \bm{A}^T \bm{y}\right)} \right)$.

We can first calculate $\frac{{\partial x_q}}{{\partial {y_p}}}$ as follows:
\begin{equation}
	\frac{\partial x_q}{\partial y_p}=\theta x_q(u_q-x_q)A_{pq}
\end{equation}

By writing the above equation into matrix form, we have:
\begin{equation}
	\frac{\partial \bm{x}}{\partial \bm{y}}={\bm{{\rm{diag}}}}(\theta\bm{x}\circ (\bm{u-x}))\bm{A}^T
\end{equation}

In the actual computation process, the matrix $\frac{\partial \bm{x}}{\partial \bm{y}}$ is not explicitly formulated for saving GPU memory. Instead, we directly formulate the derivatives $\frac{{\partial l}}{{\partial \bm{x}}}\frac{{\partial \bm{x}}}{{\partial A_{pq}}}, \frac{{\partial l}}{{\partial \bm{x}}}\frac{{\partial \bm{x}}}{{\partial b_p}}, \frac{{\partial l}}{{\partial \bm{x}}}\frac{{\partial \bm{x}}}{{\partial c_q}}, \frac{{\partial l}}{{\partial \bm{x}}}\frac{{\partial \bm{x}}}{{\partial u_q}}, \frac{{\partial l}}{{\partial \bm{x}}}\frac{{\partial \bm{x}}}{{\partial y_p}}$ in ({\ref{derivative_chain}}) as follows:
\begin{subequations}
	\begin{align}
		&\frac{{\partial l}}{{\partial \bm{x}}}\frac{{\partial \bm{x}}}{{\partial {A_{pq}}}}  = \frac{{\partial l}}{{\partial {x_q}}}\frac{{\partial {x_q}}}{{\partial {A_{pq}}}} = \frac{{\partial l}}{{\partial {x_q}}}{\theta}{x_q}\left( u_q - x_q \right){y_p} \\
		&\frac{{\partial l}}{{\partial \bm{x}}}\frac{{\partial \bm{x}}}{{\partial {b_{p}}}} = 0 \\
		&\frac{{\partial l}}{{\partial \bm{x}}}\frac{{\partial \bm{x}}}{{\partial {c_{q}}}} = \frac{{\partial l}}{{\partial {x_q}}}\frac{{\partial {x_q}}}{{\partial {c_{q}}}} = \frac{{\partial l}}{{\partial {x_q}}}{\theta x_q \left( {u_q - x_q} \right)} \\
		&\frac{{\partial l}}{{\partial \bm{x}}}\frac{{\partial \bm{x}}}{{\partial {u_{q}}}} = \frac{{\partial l}}{{\partial {x_q}}}\frac{{\partial {x_q}}}{{\partial {u_{q}}}} = \frac{{\partial l}}{{\partial {x}_q}} \frac{{x}_q - {\theta}{x}_q \left( {u}_q - {x}_q \right) \left(- c_q  - \sum\limits_{i = 1}^m {{y_i}{A_{iq}}} \right)}{{u}_q} \\
		&\frac{{\partial l}}{{\partial \bm{x}}}\frac{{\partial \bm{x}}}{{\partial {y_{p}}}} = \sum\limits_{q = 1}^n {\frac{{\partial l}}{{\partial {x_q}}}\frac{{\partial {x_q}}}{{\partial {y_{p}}}}} = \sum\limits_{q = 1}^n {\frac{{\partial l}}{{\partial {x_q}}} {\theta x_q \left( {u_q - x_q} \right)} A_{pq}}
	\end{align}
\end{subequations}

By writing the above equations into matrix form, we have:
\begin{subequations}
	\label{dldx_dxdv_matrix}
	\begin{align}
	&\frac{{\partial l}}{{\partial \bm{x}}}\frac{{\partial \bm{x}}}{{\partial \bm{A}}} = \bm{y} \left( \frac{{\partial l}}{{\partial \bm{x}}}\circ {\theta}\bm{x} \circ \left( \bm{u} - \bm{x} \right)\right)^T \\
	&\frac{{\partial l}}{{\partial \bm{x}}}\frac{{\partial \bm{x}}}{{\partial \bm{b}}} = \bm{0} \\
	&\frac{{\partial l}}{{\partial \bm{x}}}\frac{{\partial \bm{x}}}{{\partial \bm{c}}} = \frac{{\partial l}}{{\partial \bm{x}}}\circ {\theta}\bm{x} \circ \left( \bm{u} - \bm{x} \right) \\
	&\frac{{\partial l}}{{\partial \bm{x}}}\frac{{\partial \bm{x}}}{{\partial \bm{u}}} = \frac{{\partial l}}{{\partial \bm{x}}}\circ \left(\frac{\bm{x} - {\theta}\bm{x} \circ \left( \bm{u} - \bm{x} \right) \circ \left(-\bm{c} - \bm{A}^T \bm{y}\right)}{\bm{u}}\right) \\
	&\frac{{\partial l}}{{\partial \bm{x}}}\frac{{\partial \bm{x}}}{{\partial \bm{y}}} = \bm{A} \left( \frac{{\partial l}}{{\partial \bm{x}}}\circ {\theta}\bm{x} \circ \left( \bm{u} - \bm{x} \right)\right)
	\end{align}
\end{subequations}

We then calculate $\frac{{\partial {h}_p}}{{\partial {y}_q}}$ as follows:
\begin{equation}
	\frac{{\partial {h_p}}}{{\partial {y_q}}} = \sum\limits_{j = 1}^n {{A_{pj}} \theta x_j \left(u_j - x_j\right) {A_{qj}}} 
\end{equation}

By writing the above equations into matrix form, we have:
\begin{equation}
	\frac{{\partial \bm{h}}}{{\partial \bm{y}}} = \bm{A} \bm{{\rm{diag}}} \left(\theta \bm{x} \circ \left(\bm{u} - \bm{x}\right)\right) \bm{A}^T
\end{equation}
where $\bm{{\rm{diag}}} \left(\cdot\right)$ maps a vector to its corresponding diagonal matrix. When $\bm{x}$ is the optimal solution, we have $\bm{0} < \bm{x} < \bm{u}$. Therefore, $\bm{A} \bm{{\rm{diag}}} \left(\theta \bm{x} \circ \left(\bm{u} - \bm{x}\right)\right) \bm{A}^T$ is a positive semi-definite matrix.

Here, we denote $\left( {\frac{{\partial l}}{{\partial \bm{x}}}\frac{{\partial \bm{x}}}{{\partial \bm{y}}} + \frac{{\partial l}}{{\partial \bm{y}}}} \right){\left( {\frac{{\partial \bm{h}}}{{\partial \bm{y}}}} \right)^{ - 1}}$ as $\frac{{\partial l}}{{\partial \bm{h}}}$. After we finish the calculation of $\frac{{\partial l}}{{\partial \bm{h}}}$ by conjugate gradient method, we can calculate $\frac{{\partial l}}{{\partial \bm{h}}}\frac{{\partial \bm{h}}}{{\partial A_{pq}}}, \frac{{\partial l}}{{\partial \bm{h}}}\frac{{\partial \bm{h}}}{{\partial b_p}}, \frac{{\partial l}}{{\partial \bm{h}}}\frac{{\partial \bm{h}}}{{\partial c_q}}, \frac{{\partial l}}{{\partial \bm{h}}}\frac{{\partial \bm{h}}}{{\partial u_q}}$ as follows:
\begin{subequations}
	\begin{align}
		&\frac{{\partial l}}{{\partial \bm{h}}}\frac{{\partial \bm{h}}}{{\partial {A_{pq}}}} = \sum\limits_{i = 1}^m {\frac{{\partial l}}{{\partial {h_i}}}\frac{{\partial {h_i}}}{{\partial {A_{pq}}}}} = \frac{{\partial l}}{{\partial {h_p}}} x_q + {y_p}\left( {\sum\limits_{i = 1}^m {\frac{{\partial l}}{{\partial {h_i}}}{A_{iq}}} } \right) \theta x_q \left(u_q - x_q\right) \\
		&\frac{{\partial l}}{{\partial \bm{h}}}\frac{{\partial \bm{h}}}{{\partial {b_{p}}}} = \frac{{\partial l}}{{\partial {h}_p}}\frac{{\partial {h}_p}}{{\partial {b_{p}}}} = - \frac{{\partial l}}{{\partial {h}_p}} \\
		&\frac{{\partial l}}{{\partial \bm{h}}}\frac{{\partial \bm{h}}}{{\partial {c_{q}}}} = \sum\limits_{i = 1}^m {\frac{{\partial l}}{{\partial {h}_i}}\frac{{\partial {h}_i}}{{\partial {c_{q}}}}} = \left(\sum\limits_{i = 1}^m {\frac{{\partial l}}{{\partial {h_i}}} A_{iq} }\right) {\theta x_q \left( {u_q - x_q} \right)} \\
		&\frac{{\partial l}}{{\partial \bm{h}}}\frac{{\partial \bm{h}}}{{\partial {u_{q}}}} = \sum\limits_{i = 1}^m {\frac{{\partial l}}{{\partial {h}_i}}\frac{{\partial {h}_i}}{{\partial {u_{q}}}}} = \left(\sum\limits_{i = 1}^m { \frac{{\partial l}}{{\partial {h_i}}} A_{iq} }\right){ \frac{x_q - \theta x_q \left( {u_q - x_q} \right) \left(- {c}_q  - \sum\limits_{i = 1}^m {{y_i}{A_{iq}}} \right)}{u_q}}
	\end{align}
\end{subequations}

By writing the above equations into matrix form, we have:
\begin{subequations}
	\label{dldh_dhdv_matrix}
	\begin{align}
		&\frac{{\partial l}}{{\partial \bm{h}}}\frac{{\partial \bm{h}}}{{\partial \bm{A}}} = \frac{{\partial l}}{{\partial \bm{h}}} \bm{x}^T + \bm{y} \left( \left(\bm{A}^T \frac{{\partial l}}{{\partial \bm{h}}} \right) \circ \left({\theta}\bm{x} \circ \left( \bm{u} - \bm{x} \right)\right)\right)^T \\
		&\frac{{\partial l}}{{\partial \bm{h}}}\frac{{\partial \bm{h}}}{{\partial \bm{b}}} = - \frac{{\partial l}}{{\partial \bm{h}}} \\
		&\frac{{\partial l}}{{\partial \bm{h}}}\frac{{\partial \bm{h}}}{{\partial \bm{c}}} = \left(\bm{A}^T \frac{{\partial l}}{{\partial \bm{h}}} \right) \circ \left({\theta}\bm{x} \circ \left( \bm{u} - \bm{x} \right)\right) \\
		&\frac{{\partial l}}{{\partial \bm{h}}}\frac{{\partial \bm{h}}}{{\partial \bm{u}}} = \left(\bm{A}^T \frac{{\partial l}}{{\partial \bm{h}}} \right) \circ \left(\frac{\bm{x} - {\theta}\bm{x} \circ \left( \bm{u} - \bm{x} \right) \circ \left(- \bm{c} - \bm{A}^T \bm{y}\right)}{\bm{u}}\right)
	\end{align}
\end{subequations}

Finally, by substituting equations ({\ref{dldx_dxdv_matrix}}) and ({\ref{dldh_dhdv_matrix}}) into ({\ref{derivative_chain}}), we can obtain the corresponding gradient as follows:
\begin{subequations}
	\label{dldv}
	\begin{align}
		&\frac{{\partial l}}{{\partial \bm{A}}} = \bm{y} \left( \left(\frac{{\partial l}}{{\partial \bm{x}}} - \bm{A}^T \frac{{\partial l}}{{\partial \bm{h}}} \right) \circ \left({\theta}\bm{x} \circ \left( \bm{u} - \bm{x} \right)\right)\right)^T - \frac{{\partial l}}{{\partial \bm{h}}} \bm{x}^T \\
		&\frac{{\partial l}}{{\partial \bm{b}}} = \frac{{\partial l}}{{\partial \bm{h}}} \\
		&\frac{{\partial l}}{{\partial \bm{c}}} = \left( \frac{{\partial l}}{{\partial \bm{x}}}\circ {\theta}\bm{x} \circ \left( \bm{u} - \bm{x} \right)\right) - \left(\bm{A}^T \frac{{\partial l}}{{\partial \bm{h}}} \right) \circ \left({\theta}\bm{x} \circ \left( \bm{u} - \bm{x} \right)\right) \\
		&\frac{{\partial l}}{{\partial \bm{u}}} = \left(\frac{{\partial l}}{{\partial \bm{x}}} - \bm{A}^T \frac{{\partial l}}{{\partial \bm{h}}} \right) \circ \left(\frac{\bm{x} - {\theta}\bm{x} \circ \left( \bm{u} - \bm{x} \right) \circ \left(- \bm{c} - \bm{A}^T \bm{y}\right)}{\bm{u}}\right)
	\end{align}
\end{subequations}

\subsection{Experimental details about constrained traveling salesman problem} \label{appendix.a5}
Here, we first provide the mathematical formulation of TSP with starting and ending cities constraint (TSP-StartEnd) and TSP with priority constraint (TSP-Priority).
We first focus on TSP-StartEnd, which can be modeled as follows:
\begin{mini!}[2]
	{_{\bm{X} \in \left\{ {0, 1} \right\}^{n \times n}}}
	{\label{tsp_se_obj} \sum\limits_{i = 1}^n {\sum\limits_{j = 1}^n {{D_{i,j}}\sum\limits_{k = 1}^{n - 1} {{X_{i,k}}{X_{j,k + 1}}} } } }
	{\label{tsp_se}}
	{}
	\addConstraint{ X_{s, 1} = 1, X_{e, n} = 1}{  \label{cons_tsp_se_1}}{}
	\addConstraint{ \bm{X}^T \bm{1}_n = \bm{1}_n, \bm{X} \bm{1}_n = \bm{1}_n}{\label{cons_tsp_se_2}}{}
\end{mini!}
where $X_{i,j} = 1$ means city $i$ is the $j$-th visited city in a tour, $D_{i, j}$ refers to the distance between city $i$ and city $j$.

On the basis of problem ({\ref{tsp_se}}), the TSP-Priority problem can be obtained by introducing the following priority constraints:
\begin{equation}
	\sum\limits_{j = 1}^{m + 1} {{X_{p,j}}}  = 1
\end{equation}
where $p$ is the city that needs to be visited in the first $m$ steps.

Due to the focus of this experiment on comparing the performance of each satisfiability layer, designing a better network structure is beyond the scope of this paper. Therefore, we directly used the SOTA network structure in solving TSP, which consists of a three-layers Transformer, followed by a three-layer MLP with ReLU activation {\cite{wang2023linsatnet}}. The hidden sizes are all set to 256 and the number of multi-head attention is set to 8. All the training and test data is consisted of 20 nodes that are uniformly sampled from a unit square. For all satisfiability layers, the constraints are the continuous relaxation of the original TSP problem and all the common settings are the same as follows. The learning rate is set to 10\textsuperscript{-4}. The batch size is set to 1024. When we stack constraints into block diagonal form to exploit parallelism, there are about 40,000 rows and 400,000 columns in the whole matrix. The training epoch number is set to 50. The constraint tolerance is set to 10\textsuperscript{-3}. For GLinSAT, the initial estimate of Lipschitz constant is set to $\theta$, the initial estimate of the dual variable is set to zero vector, the numerical precision is set to $10 \epsilon_{\rm{machine}}$, where $\epsilon_{\rm{machine}}$ is the machine epsilon. In each epoch, we generate 256000 random cases as the training set. In the training stage, the objective function in ({\ref{tsp_se_obj}}) is used as the loss function. In the validation stage, we use two kinds of post-processing methods. The first one is rounding. The second one is beam search where the width of the beam is set to 2048. All the experiments are conducted on a computer with a Intel(R) Xeon(R) Platinum 8360H CPU and a NVIDIA Tesla A100 GPU with 80GB memory through Pytorch 2.2.

Here, we additionally show the training performance when $1 / \theta$ is set to 10\textsuperscript{-2} as follows:
\begingroup
\renewcommand{\arraystretch}{1.2}
\begin{table}[h]
	\small
	\centering
	\caption{Average allocated GPU memory and solution time of different satisfiability layers during batch processing of projection and backpropagation when $1 / \theta$ is set to 10\textsuperscript{-2} in TSP training phase}
	\begin{tabularx}{\textwidth}{M{0.234\textwidth}M{0.0426\textwidth}CM{0.0426\textwidth}CM{0.0426\textwidth}CM{0.0426\textwidth}C}
			\toprule
			& \multicolumn{4}{c}{TSP-StartEnd}       & \multicolumn{4}{c}{TSP-Priority} \\
			\cmidrule(lr){2-5} \cmidrule(lr){6-9}
			& \multicolumn{2}{c}{GPU Mem./MB} & \multicolumn{2}{c}{Time/s} & \multicolumn{2}{c}{GPU Mem./MB} & \multicolumn{2}{c}{Time/s} \\
			\cmidrule(lr){2-3} \cmidrule(lr){4-5} \cmidrule(lr){6-7} \cmidrule(lr){8-9}
			Satisfiability layer & Proj. & Backprop.  & Proj. & Backprop.  & Proj. & Backprop.  & Proj. & Backprop.   \\
			\midrule
			CvxpyLayers   & \---\---     & \---\---    & 107.3     & 31.12     & \---\---     & \---\---     & 107.4     & 30.22 \\
			OptNet  & 14305     & 5005    & 18.89    & 0.930    & 14333    & 5005    & 20.40    & 1.125 \\
			LinSAT-Dense-100 & 14977    & 181.2    & \textbf{0.278}    & 0.417    & 15009    & 180.9    & \textbf{0.277}    & 0.419 \\
			LinSAT-Dense-500 & 74108    & 181.2    & 1.339    & 1.930    & 74272    & 180.9    & 1.318    & 1.932 \\
			GLinSAT-Dense-Explicit & 11095    & \textbf{4.852}    & 0.976    & 0.552    & 11155    & \textbf{4.857}    & 0.965    & 0.555 \\
			\cellcolor{gray!20}GLinSAT-Dense-Implicit & \cellcolor{gray!20}{\textbf{13.34}}    & \cellcolor{gray!20}{53.23}    & \cellcolor{gray!20}{0.888}    & \cellcolor{gray!20}{\textbf{0.072}}    & \cellcolor{gray!20}{\textbf{13.36}}    & \cellcolor{gray!20}{53.22}    & \cellcolor{gray!20}{1.096}    & \cellcolor{gray!20}{\textbf{0.078}} \\
			LinSAT-Sparse-100 & \---\---    & \---\---    & \---\---    & \---\---    & \---\---    & \---\---    & \---\---    & \---\--- \\
			LinSAT-Sparse-500 & \---\---    & \---\---    & \---\---    & \---\---    & \---\---    & \---\---    & \---\---    & \---\--- \\
			GLinSAT-Sparse-Explicit & 7085    & 5.051    & 1.552    & 0.774    & 7007    & 5.519    & 1.508    & 0.812 \\
			GLinSAT-Sparse-Implicit & 62.95    & 24.71    & 1.355    & 0.078    & 63.27    & 24.51    & 1.721    & 0.083 \\
			\bottomrule
		\end{tabularx}%
	\begin{tablenotes}
			\item[a] Note: LinSAT-(Dense/Sparse)-(100/500) means that LinSAT is used with dense/sparse matrix and max iteration number is set to 100/500. GLinSAT-(Dense/Sparse)-(Explicit/Implicit) means that GLinSAT is used with dense/sparse matrix and backpropagation is performed using automatic differential/implicit differential. The GPU memory used by CvxpyLayers is not counted since CvxpyLayers is CPU-based.
			\item[b] Note: The output of LinSAT-Sparse when $1/\theta=0.01$ is not a real number so that the results are not shown.
		\end{tablenotes}
	\label{tab_performance_tsp_2}%
\end{table}%
\endgroup

From Table {\ref{tab_performance_tsp_2}}, we can see GLinSAT-Dense-Implicit is the most memory efficient satisfiability layer among all these layers. As to the solution time, we can find that GLinSAT is slightly slower than LinSAT-Dense-100. \textbf{The reason LinSAT has a fast calculation speed is that the algorithm terminates due to reaching the maximum number of iterations rather than because of convergence.} Actually we find that LinSAT often reports warnings like "non-zero constraint violation within max iterations", which indicates the algorithm has not converged. After we increase the max iteration number to 500, the number of warnings has decreased, but there are still some warnings that indicate the algorithm has not converged. When we set maximum iteration number to $+\infty$, the algorithm progress will stuck. Compared with LinSAT, our proposed GLinSAT is more reliable and is guaranteed to converge. Table {\ref{tab_result_tsp_2}} shows the corresponding validation results about the mean tour length and feasibility ratio. From Table {\ref{tab_result_tsp_2}}, we can see that the performance of GLinSAT is superior to that of LinSAT.

\begingroup
\renewcommand{\arraystretch}{1.0}
\begin{table}[htbp]
	\small
	\centering
	\caption{Mean tour length and feasibility ratio obtained from using different $1 / \theta$ and post-processing methods in TSP validation stage}
	\begin{tabularx}{\textwidth}{M{0.234\textwidth}M{0.0625\textwidth}M{0.0625\textwidth}M{0.0625\textwidth}M{0.0625\textwidth}M{0.0625\textwidth}M{0.0625\textwidth}M{0.0625\textwidth}M{0.0625\textwidth}}
		\toprule
		& \multicolumn{4}{c}{TSP-StartEnd}       & \multicolumn{4}{c}{TSP-Priority} \\
		\cmidrule(lr){2-5} \cmidrule(lr){6-9}
		& \multicolumn{2}{c}{\makecell{Rounding \\with $1 / \theta$ = 10\textsuperscript{-3} }} & \multicolumn{2}{c}{\makecell{Beamsearch \\with $1 / \theta = 0.1$ }} & \multicolumn{2}{c}{\makecell{Rounding \\with $1 / \theta $ = 10\textsuperscript{-3} }} & \multicolumn{2}{c}{\makecell{Beamsearch \\with $1 / \theta = 0.1$ }} \\
		\cmidrule(lr){2-3} \cmidrule(lr){4-5} \cmidrule(lr){6-7} \cmidrule(lr){8-9}
		Layer  & Mean Length & Feas. Ratio & Mean Length & Feas. Ratio & Mean Length & Feas. Ratio & Mean Length & Feas. Ratio \\
		\midrule
		LinSAT-Dense-100 & \---\---    & \---\---    & 8.090    & \textbf{100\%}    & \---\---    & \---\---    & 7.031    & \textbf{100\%} \\
		LinSAT-Dense-500 & \---\---    & \---\---    & 3.873    & \textbf{100\%}    & \---\---    & \---\---    & 4.011    & \textbf{100\%} \\
		GLinSAT-Dense-Explicit & 4.030    & 92.7\%    & 3.869    & \textbf{100\%}   & 4.171    & 91.5\%    & 3.983    & \textbf{100\%} \\
		\cellcolor{gray!20}GLinSAT-Dense-Implicit & \cellcolor{gray!20}\textbf{3.930}    & \cellcolor{gray!20}94.1\%    & \cellcolor{gray!20}\textbf{3.790}    & \cellcolor{gray!20}\textbf{100\%}    & \cellcolor{gray!20}\textbf{4.053}    & \cellcolor{gray!20}\textbf{94.6\%}    & \cellcolor{gray!20}\textbf{3.924}    & \cellcolor{gray!20}\textbf{100\%} \\
		LinSAT-Sparse-100 & \---\---    & \---\---    & \---\---    & \---\---    & \---\---    & \---\---   & \---\---    & \---\--- \\
		LinSAT-Sparse-500 & \---\---    & \---\---    & \---\---    & \---\---    & \---\---    & \---\---    & \---\---    & \---\--- \\
		GLinSAT-Sparse-Explicit & 4.022    & 92.6\%    & 3.859    & \textbf{100\%}    & 4.078    & 92.6\%    & 3.935    & \textbf{100\%} \\
		GLinSAT-Sparse-Implicit & 3.926    & \textbf{94.8\%}    & 3.803    & \textbf{100\%}    & 4.073    & 93.4\%    & 3.933    & \textbf{100\%} \\
		\bottomrule
	\end{tabularx}%
	\begin{tablenotes}
		\item[a] Note: The output of LinSAT-Sparse when $\frac{1}{\theta}$ = 10\textsuperscript{-2} is not a real number so that we cannot obtain any trained model. The output of LinSAT-Dense when $\frac{1}{\theta}$ = 10\textsuperscript{-3} is not a real number so that the results are not shown.
	\end{tablenotes}
	\label{tab_result_tsp_2}%
\end{table}
\endgroup

\subsection{Experimental details about partial graph matching with outliers} \label{appendix.a6}
Here, we first provide the mathematical formulation of partial graph matching with outliers. Denote $m, n$ as the number of nodes of two graphs respectively. The partial graph matching problem with $p$ inliers can be expressed as follows:
\begin{subequations}
	\label{cons_matching}
	\begin{gather}
		\label{cons_matching_1}
		\bm{X}^T \bm{1}_m \leq \bm{1}_n \\
		\label{cons_matching_2}
		\bm{X}\bm{1}_n \leq \bm{1}_m \\
		\bm{1}_m^T \bm{X} \bm{1}_n = p \\
		\bm{X} \in \left\{0, 1\right\}^{m \times n}
	\end{gather}
\end{subequations}
where $X_{i,j} = 1$ means the $i$-th node in the left graph matches the $j$-th node in the right graph.

When we use GLinSAT as the satisfiability layer, it is necessary to canonize the original inequality constraints. By introducing bounded slack variables into equations ({\ref{cons_matching_1}}) and ({\ref{cons_matching_2}}), we can reformulate constraints ({\ref{cons_matching}}) into the standard form as follows:
\begin{subequations}
	\label{cons_matching_reform}
	\begin{gather}
		\label{cons_matching_reform_1}
		\bm{X}^T \bm{1}_m + \bm{s}_n = \bm{1}_n \\
		\label{cons_matching_reform_2}
		\bm{X}\bm{1}_n + \bm{t}_m = \bm{1}_m \\
		\bm{1}_m^T \bm{X} \bm{1}_n = p \\
		\bm{X} \in \left\{0, 1\right\}^{m \times n}
	\end{gather}
\end{subequations}
where $\bm{0}_n \leq \bm{s}_n \leq \bm{1}_n, \bm{0}_m \leq \bm{t}_m \leq \bm{1}_m$.

In the training stage, we follow the experimental codes in Ref. {\cite{wang2023linsatnet}} where the neural networks are trained on the basis of a pretrained SOTA graph matching NGMv2 model {\cite{wang2021neural}} named "pretrained\_params\_vgg16\_ngmv2\_afat-i\_voc". Different satisfiability layers are used to make the outputs satisfy the continuous relaxation of constraints ({\ref{cons_matching}}).

It is noteworthy that since the sizes of graphs differ a lot in one batch, we stack constraints into block diagonal forms in LinSAT and GLinSAT to exploit parallelism of the GPU. However, it is difficult for CvxpyLayers and OptNet to directly handle large block diagonal matrices since disciplined parameterized programming compilation and matrix factorization of large matrices will cost a large amount of time. Therefore, we can only use a sequential way to handle batched graphs with different sizes for CvxpyLayers and OptNet. The batch size is set to 128 across all the experiments. When we stack constraints into block diagonal form to exploit parallelism, there are about 2,500 rows and 13,000 columns in the whole matrix. The constraint tolerance is set to 10\textsuperscript{-3}. In the training stage, binary cross entropy loss is used as the loss function. For experiments about OptNet and GLinSAT-Explicit, we use double-precision floating-point numbers during projection. If single-precision floating-point numbers are used, OptNet will encounter numerical issues in its forward pass while reporting warnings like "Returning an inaccurate and potentially incorrect solution". GLinSAT-Explicit will not encounter numerical issues in its forward pass. However, sometimes the gradient calculated by auto differential may be not a real number. We believe the problem is that single-precision floating-point numbers amplify the cumulative error of backpropagation. When we use double-precision floating-point numbers, everything works fine. As a result, we use single-precision floating-point numbers on all the other layers except OptNet and GLinSAT-Explicit.

In the validation stage, we use Hungarian algorithm and greedy strategy for post-processing {\cite{wang2023linsatnet}}. We can regard the cost of matching a pair of nodes as the outputs of satisfiability layers, then use Hungarian algorithm to obtain a maximum matching. Finally, we can use greedy strategy to preserve pairs with $p$-highest matching scores and obtain the solution. 

All the experiments are conducted on a computer with a Intel(R) Xeon(R) Platinum 8360H CPU and a NVIDIA Tesla A100 GPU with 80GB memory through Pytorch 2.2. For GLinSAT, the initial estimate of Lipschitz constant is set to $\theta$, the initial estimate of the dual variable is set to zero vector, the numerical precision is set to $10 \epsilon_{\rm{machine}}$, where $\epsilon_{\rm{machine}}$ is the machine epsilon. Here, we show the average memory usage and the solution time of different satisfiability layers in Table {\ref{tab_performance_match}}.
\begingroup
\renewcommand{\arraystretch}{1.05}
\begin{table}[h]
	\small
	\centering
	\caption{Average allocated GPU memory and solution time of different satisfiability layers during batch processing of projection and backpropagation in the training phase of partial graph matching}
	\begin{tabularx}{\textwidth}{M{0.234\textwidth}M{0.0426\textwidth}CM{0.0426\textwidth}CM{0.0426\textwidth}CM{0.0426\textwidth}C}
		\toprule
		& \multicolumn{4}{c}{$\frac{1}{\theta}$ = 10\textsuperscript{-1}}       & \multicolumn{4}{c}{$\frac{1}{\theta}$ = 10\textsuperscript{-2}} \\
		\cmidrule(lr){2-5} \cmidrule(lr){6-9}
		& \multicolumn{2}{c}{GPU Mem./MB} & \multicolumn{2}{c}{Time/s} & \multicolumn{2}{c}{GPU Mem./MB} & \multicolumn{2}{c}{Time/s} \\
		\cmidrule(lr){2-3} \cmidrule(lr){4-5} \cmidrule(lr){6-7} \cmidrule(lr){8-9}
		Layer & Proj. & Backprop.  & Proj. & Backprop.  & Proj. & Backprop.  & Proj. & Backprop.   \\
		\midrule
		CvxpyLayers   & \---\---     & \---\---    & 64.26     & 16.09     & \---\---     & \---\---     & 12.76     & 15.07 \\
		OptNet  & 167.3     & 826.2    & 4.290    & \textbf{3.712}    & 167.2    & 826.3    & 4.437    & 3.726 \\
		LinSAT-Dense-100\textsuperscript{*} & 23944    & 322.1    & 0.611    & 3.887    & 23944    & 322.1    & 0.611    & 3.887 \\
		LinSAT-Dense-500 & $\times$    & $\times$    & $\times$    & $\times$    & $\times$    & $\times$    & $\times$    & $\times$ \\
		GLinSAT-Dense-Explicit & 1249    & \textbf{0.997}    & 1.891    & 4.418    & 1311    & \textbf{1.052}    & 1.991    & 4.147 \\
		\cellcolor{gray!20}GLinSAT-Dense-Implicit & \cellcolor{gray!20}{129.0}    & \cellcolor{gray!20}{736.5}    & \cellcolor{gray!20}{\textbf{1.333}}    & \cellcolor{gray!20}{3.995}    & \cellcolor{gray!20}{129.0}    & \cellcolor{gray!20}{736.4}    & \cellcolor{gray!20}{\textbf{1.302}}    & \cellcolor{gray!20}{\textbf{3.515}} \\
		LinSAT-Sparse-100\textsuperscript{*} & 803.0    & 397.7    & 1.593    & 3.906    & 803.6    & 397.7    & 1.615    & 3.877 \\
		LinSAT-Sparse-500\textsuperscript{*} & 2772    & 304.3    & 2.922   & 5.534    & 2772    & 304.3    & 2.923    & 5.292 \\
		GLinSAT-Sparse-Explicit & 648.7    & 172.9    & 1.794    & 4.435    & 687.6    & 139.8    & 1.986    & 4.620 \\
		GLinSAT-Sparse-Implicit & \textbf{0.001}    & 862.1    & 1.544    & 3.884    & \textbf{0.001}    & 862.1    & 1.449   & 3.865 \\
		\bottomrule
	\end{tabularx}%
	\begin{tablenotes}
		\item[a] Note: The GPU memory used by CvxpyLayers is not counted since CvxpyLayers is CPU-based.
		\item[b] Note: The symbol "*" means the outputs of this satisfiability layer cannot meet constraints. The symbol "$\times$" means the layer leads to out-of-memory (OOM) issues.
	\end{tablenotes}
	\label{tab_performance_match}%
\end{table}%
\endgroup

In Table {\ref{tab_performance_match}}, although LinSAT-100 seems to be the fastest method, the price is that the required constraints are not satisfied at all. When we set the maximum iteration number of LinSAT to 100, almost every batch LinSAT will report warnings like "non-zero constraint violation within max iterations". When we set the maximum iteration number to 500, LinSAT-Dense will soon run out of memory while LinSAT-Sparse will still display the warning message in almost every epoch. Notably, the computational speed of LinSAT at this point has already fallen behind that of GLinSAT. If we set the maximum iteration number to $+\infty$ in LinSAT-Sparse, the algorithm will get stuck. Compared with our proposed GLinSAT, when the maximum iteration number is set to $+\infty$, the convergence of GLinSAT is guaranteed. In summary, we can conclude that our proposed GLinSAT is the fastest satisfiability layer while ensuring the outputs satisfy the linear and bounded constraints from the results shown in Table {\ref{tab_performance_match}}.

\subsection{Experimental details about predictive portfolio allocation} \label{appendix.a7}
Here, we first restate the mathematical formulation of predictive portfolio allocation. We denote $x_i$ as the predicted portfolio decision variable of asset $i$ and $\cal S$ as the preferred portfolio asset. The portfolio allocation needs to maximize the Sharpe ratio {\cite{sharpe1998sharpe}} while satisfying the following constraints:
\begin{subequations}
	\label{cons_port}
	\begin{gather}
		\label{cons_port_linear}
		\sum\limits_{i = 1}^n {{x_i}}  = 1, \sum\limits_{i \in \cal S}^{} {{x_i}} \ge q \\
		0 \leq x_i \leq 1, \forall 1 \leq i \leq n
	\end{gather}
\end{subequations}
where $q$ is a pre-defined positive constant. Following the codes in {\cite{wang2023linsatnet}}, here we set $q$ as 0.5 and set $C$ as \{AAPL, MSFT, AMZN, TSLA, GOOGL\}. We also use the first 120-day historical data to train a neural network that could maximize the Sharpe ratio for the future 120 days where StemGNN {\cite{cao2020spectral}} are used as the network backbone to extract the features. 

When we use GLinSAT as the satisfiability layer, it is necessary to canonize the original inequality constraints. By introducing bounded slack variables into constraints ({\ref{cons_port_linear}}), we can reformulate constraints ({\ref{cons_port}}) into the standard form as follows:
\begin{subequations}
	\begin{gather}
		\sum\limits_{i = 1}^n {{x_i}}  = 1, \sum\limits_{i \in \cal S}^{} {{x_i}} - w = q \\
		0 \leq w \leq \left| \cal S \right| - q, 0 \leq x_i \leq 1, \forall 1 \leq i \leq n
	\end{gather}
\end{subequations}
where $\left| \cal S \right|$ refers to the number of elements in the preference set $\cal S$.

To conduct fair comparison, we train 50 epochs with a batch size of 128, a learning rate of 10\textsuperscript{-5} and a constraint tolerance of 10\textsuperscript{-3} across all satisfiability layers. When we stack constraints into block diagonal form to exploit parallelism, there are about 250 rows and 60,000 columns in the whole matrix. In the training stage, a weighted sum of prediction MSE error on future asset prices and the opposite of Sharpe ratio is used as the loss function. All the experiments are conducted on a computer with a Intel(R) Xeon(R) Platinum 8360H CPU and a NVIDIA Tesla A100 GPU with 80GB memory through Pytorch 2.2. For GLinSAT, the initial estimate of Lipschitz constant is set to $\theta$, the initial estimate of the dual variable is set to zero vector, the numerical precision is set to $10 \epsilon_{\rm{machine}}$, where $\epsilon_{\rm{machine}}$ is the machine epsilon. In the main text, we have reported the mean Sharpe ratio of each method in Table {\ref{tab_result_port}}. Here, we provide the results related to the training performance in Table {\ref{tab_performance_port}}. From Table {\ref{tab_performance_port}}, we can see that the total memory usage is similar between each variant of LinSAT and GLinSAT since there are only two constraints in the original problem. The projection time of LinSAT and GLinSAT is significantly less than that of CvxpyLayers and OptNet while GLinSAT-Dense-Implicit use the shortest total calculation time among all satisfiability layers.

\begingroup
\renewcommand{\arraystretch}{1.1}
\begin{table}[h]
	\small
	\centering
	\caption{Average allocated GPU memory and solution time of different satisfiability layers during batch processing of projection and backpropagation in the training phase of predictive portfolio allocation}
	\begin{tabularx}{\textwidth}{M{0.234\textwidth}M{0.0426\textwidth}CM{0.0426\textwidth}CM{0.0426\textwidth}CM{0.0426\textwidth}C}
			\toprule
			& \multicolumn{4}{c}{$\frac{1}{\theta}$ = 10\textsuperscript{-1}}       & \multicolumn{4}{c}{$\frac{1}{\theta}$ = 10\textsuperscript{-2}} \\
			\cmidrule(lr){2-5} \cmidrule(lr){6-9}
			& \multicolumn{2}{c}{GPU Mem./MB} & \multicolumn{2}{c}{Time/s} & \multicolumn{2}{c}{GPU Mem./MB} & \multicolumn{2}{c}{Time/s} \\
			\cmidrule(lr){2-3} \cmidrule(lr){4-5} \cmidrule(lr){6-7} \cmidrule(lr){8-9}
			Layer & Proj. & Backprop.  & Proj. & Backprop.  & Proj. & Backprop.  & Proj. & Backprop.   \\
			\midrule
			CvxpyLayers   & \---\---     & \---\---    & 12.85     & 1.476     & \---\---     & \---\---     & 12.27     & 1.423 \\
			OptNet  & 2012     & 765.0    & 5.599    & 1.035    & 2012    & 765.0    & 4.606    & 1.279 \\
			LinSAT-Dense-100 & 81.34    & 434.9    & 0.091    & 0.393    & 105.8    & 410.5    & 0.127    & 0.413 \\
			LinSAT-Dense-500 & 81.34    & 434.9    & 0.266    & 0.386    & 136.6    & 385.6    & 0.315    & 0.413 \\
			GLinSAT-Dense-Explicit & 112.9   & 402.6    & 0.143    & 0.376    & 107.9    & 407.2   & 0.148    & 0.377 \\
			\cellcolor{gray!20}GLinSAT-Dense-Implicit & \cellcolor{gray!20}{\textbf{1.565}}    & \cellcolor{gray!20}{514.1}    & \cellcolor{gray!20}{\textbf{0.090}}    & \cellcolor{gray!20}{0.320}    & \cellcolor{gray!20}{\textbf{1.565}}    & \cellcolor{gray!20}{513.6}    & \cellcolor{gray!20}{\textbf{0.090}}    & \cellcolor{gray!20}{0.322} \\
			LinSAT-Sparse-100 & 223.6    & \textbf{295.0}    & 0.113    & 0.400    & 290.7    & 227.6    & 0.161    & 0.421 \\
			LinSAT-Sparse-500 & 223.6    & \textbf{295.0}    & 0.114    & 0.398    & 371.1    & \textbf{224.1}    & 0.225    & 0.442 \\
			GLinSAT-Sparse-Explicit & 78.24    & 437.7    & 0.211    & 0.391    & 74.80    & 440.8    & 0.213    & 0.392 \\
			GLinSAT-Sparse-Implicit & 4.501    & 512.0    & 0.143    & \textbf{0.316}    & 4.501    & 511.7    & 0.139    & \textbf{0.315} \\
			\bottomrule
		\end{tabularx}%
	\label{tab_performance_port}%
\end{table}%
\endgroup

\subsection{Experimental details about power system unit commitment} \label{appendix.a8}
In this section, we carry out experiments on power system unit commitment where the data comes from a real provincial power system. 

We first briefly introduce the unit commitment problem. Unit commitment problem is a core optimization problem in power system operation and planning.  It mainly involves deciding how to most economically and safely arrange the startup-shutdown status and output power of generators while ensuring constraints related to equipment and grid operation can be satisfied. Generally, constraints can be divided into soft constraints and hard constraints. In general, constraints directly related to generators are often regarded as hard constraints, e.g. the generator minimum up-time and down-time constraints. Constraints related to the section power and load balance are usually regarded as soft constraints. For these constraints, we often penalize the corresponding violation in the objective {\cite{wu2021novel}}.

Next, we will use the common three-binary formulation {\cite{knueven2020mixed}} to model the unit commitment problem. Let $T$ denote the total number of time steps that considered in unit commitment problem. $G$ is the total number of generators. We denote ${\cal T} = \left\{1, \cdots, T\right\}$ and ${\cal G} = \left\{1, \cdots, G\right\}$. Let $u_g(t)$ denote whether the generator $g$ is on at time $t$, $v_g(t)$ denote whether the generator is turned on at time $t$, and $w_g(t)$ denote if the generator is turned off at time $t$. Then, $u_g\left(t\right), v_g\left(t\right), w_g\left(t\right)$ satisfy the following logical constraints:
\begin{equation}
	\label{cons_logical}
	u_g\left(t\right) - u_g\left(t - 1\right) = v_g\left(t\right) - w_g\left(t\right), t  \in \cal T
\end{equation}
where $T$ is the total number of time steps that considered in the unit commitment problem.

Units need to satisfy minimum up-time constraints and down-time constraints as follows:
\begin{subequations}
	\label{cons_up_down}
	\begin{gather}
		\sum\limits_{i = t - U{T_g} + 1}^t {{v_g}\left( i \right)}  \le {u_g}\left( t \right), g \in {\cal G}, t = U{T_g}, \cdots T \\
		\sum\limits_{i = t - D{T_g} + 1}^t {{w_g}\left( i \right)}  \le 1 - {u_g}\left( t \right), g \in {\cal G}, t = D{T_g}, \cdots T
	\end{gather}
\end{subequations}
where $UT_g, DT_g$ are the minimum up-time and minimum down-time for generator $g$ respectively.

Let $p_g\left(t\right)$ denote the power produced by generator $g$ at time $t$. Let $\underline{p_g}, \overline{p_g}$ denote the lower bound and upper bound of generator $g$. Then, we have constraints related to the bound of generator output as follows:
\begin{equation}
	\underline{p_g}u_g\left(t\right) \leq p_g\left(t\right) \leq \overline{p_g}u_g\left(t\right), g \in {\cal G}, t \in {\cal T}
\end{equation}

We also need to consider the ramping capability of each generator. The ramping constraints can be formulated as follows:
\begin{subequations}
	\begin{gather}
		{p_g}\left( t \right) - {p_g}\left( {t - 1} \right) \le -RU_g v_g\left(t\right) + \left(\underline{p_g} + RU_g\right) u_g\left(t\right) - \underline{p_g}u_g\left(t - 1\right), g \in {\cal G}, t \in {\cal T} \\
		{p_g}\left( t - 1 \right) - {p_g}\left( t \right) \le -RD_g w_g\left(t\right) + \left(\underline{p_g} + RD_g\right) u_g\left(t - 1\right) - \underline{p_g}u_g\left(t\right), g \in {\cal G}, t \in {\cal T}
	\end{gather}
\end{subequations}
where $RU_g, RD_g$ denote the ramp-up rate and ramp-down rate of generator $g$.

The load generation balance constraint is modeled as the following soft constraint form:
\begin{equation}
	\sum\limits_{g \in {\cal G}} {{p_g}\left( t \right)}  + {s^ + }\left( t \right) - {s^ - }\left( t \right) = \sum\limits_{d \in {\cal D}} {l_d\left( t \right)}, t \in {\cal T}
\end{equation}
where $\cal D$ is the set of all loads, $l_d\left(t\right)$ represents the $d$-th load at time $t$, $s^+\left(t\right) \geq 0, s^-\left(t\right) \geq 0$ are the non-negative slack variables at time $t$ which will be later penalized in the objective.

We denote the set of sections as ${\cal K}$. The soft constraints on section power are provided as follows:
\begin{equation}
	\label{cons_section}
	\underline {{F_k}}  \le \sum\limits_{g \in \cal G} {{H_{kg}}{p_g}\left( t \right)}  - \sum\limits_{d \in \cal D} {{H_{kd}}{l_d}\left( t \right)}  + s_k^ + \left( t \right) - s_k^ - \left( t \right) \le \overline {{F_k}}, k \in {\cal K}, t \in {\cal T}
\end{equation}
where $\underline{F_k}, \overline{F_k}$ are the lower bound and upper bound of the $k$-th section power, $H_{kg}$ is the generation shift factor that indicates the change of the $k$-th section power with respect to a change in injection at generator $g$, $H_{kg}$ is the load shift factor that indicates the change of the $k$-th section power with respect to a change in injection at load $d$, $s_k^+\left(t\right) \geq 0, s_k^-\left(t\right) \geq 0$ are the non-negative slack variables related to the $k$-th section at time $t$ which will be later penalized in the objective.

Finally, we want to minimize the system operation cost and we can obtain the optimization problem as follows:
\begin{mini!}[2]
	{}
	{\label{uc_obj} \sum\limits_{t \in {\cal T}} {\sum\limits_{g \in {\cal G}} {\left( {{c_g}{p_g}\left( t \right) + c_g^{SU}{v_g}\left( t \right)} \right)} }  + \sum\limits_{t \in \cal T} {M\left( {{s^ + }\left( t \right) + {s^ - }\left( t \right)} \right)}  + \sum\limits_{t \in \cal T} {\sum\limits_{k \in \cal K} {{M_k}\left( {s_k^ + \left( t \right) + s_k^ - \left( t \right)} \right)} }  }
	{\label{uc_prob}}
	{}
	\addConstraint{ \left(\textrm{\ref{cons_logical}}\right) - \left(\textrm{\ref{cons_section}}\right)}{  \nonumber}{}
	\addConstraint{ u_g\left(t\right) \in \left\{0, 1\right\}, v_g\left(t\right) \in \left\{0, 1\right\}, w_g\left(t\right) \in \left\{0, 1\right\}, g \in {\cal G}, t \in {\cal T} }{}{}
	\addConstraint{s_k^+\left(t\right) \geq 0, s_k^-\left(t\right) \geq 0, t \in {\cal T}}{}{}
	\addConstraint{s^+\left(t\right) \geq 0, s^-\left(t\right) \geq 0, k \in {\cal K}, t \in {\cal T}}{}{}
\end{mini!}
where $M, M_k$ are pre-defined penalty coefficients, $c_g$ is the generator cost coefficient, $c_g^{SU}$ is the generator start-up cost.

The power system we use in this article contains about 360 units. We set the total number of time steps $T$ as 96 where the interval between two adjacent time steps is set to 15 minutes. More than 1,400 sections need to be considered in the unit commitment problem. The penalty coefficient for load imbalance is set to 10\textsuperscript{11} and the penalty coefficient for section power violation is set to 10\textsuperscript{7}. Based on the one-year load data, we solve the unit commitment problem via Gurobi within a 0.1\% optimality gap. We can then obtain the optimal unit states for further supervised learning. 

In supervised learning, we want to predict the optimal value of $u_g$ as accurately as possible so that we can fix these binary variables and quickly obtain a high-quality solution from solving a linear programming problem. Therefore, we should require the predicted variables $u_g$ to satisfy the hard constraints ({\ref{cons_logical}}) and ({\ref{cons_up_down}}), namely the logical constraint and the minimum up-time and down-time constraints. In addition, $u_g\left(t\right), v_g\left(t\right), w_g\left(t\right)$ should be in the range between 0 and 1. We use satisfiability layers to ensure the above constraints can be satisfied.

When we use GLinSAT as the satisfiability layer, it is necessary to canonize the original inequality constraints. By introducing bounded slack variables into constraints ({\ref{cons_port_linear}}), we can reformulate constraints ({\ref{cons_port}}) into the standard form as follows:
\begin{subequations}
	\label{cons_eq_up_down}
	\begin{gather}
		\sum\limits_{i = t - U{T_g} + 1}^t {{v_g}\left( i \right)} + sv_g\left(t\right) = {u_g}\left( t \right), g \in {\cal G}, t = U{T_g}, \cdots T \\
		\sum\limits_{i = t - D{T_g} + 1}^t {{w_g}\left( i \right)} + sw_g\left(t\right)  = 1 - {u_g}\left( t \right), g \in {\cal G}, t = D{T_g}, \cdots T
	\end{gather}
\end{subequations}

Since we have $u_g\left(t\right) \in \left[0, 1\right], v_g\left(t\right) \in \left[0, 1\right], w_g\left(t\right) \in \left[0, 1\right]$, the value range of variables $sv_g\left(t\right)$ and $sw_g\left(t\right)$ can be quickly inferred, which is exactly $sv_g\left(t\right) \in \left[0, 1\right], sw_g\left(t\right) \in \left[0, 1\right]$.

We use a MLP-based neural network to learn the optimal unit states. The loads at each time-step are forwarded to a 2-layer MLP where the hidden sizes are set to 32. Then, we concatenate the embeddings at all time steps, and forward the concatenated embedding to a 1-layer MLP where the hidden size is set to 3072. Finally, we use a 2-layer MLP to read out the optimal unit states where the hidden size is set to 512. We use ReLU as the activation function in the hidden layers. We set the learning rate to 10\textsuperscript{-3} and train neural networks for 100 epochs. We use 70\% of one-year data as the training set and the other as the validation set. In the training stage, binary cross entropy loss is used as the loss function. For all satisfiability layers, the batch size is set to 16 and the regularization parameter $\frac{1}{\theta}$ is set to 0.1. It is worth noting that although the batch size is not that large, the scale of the optimization problem in each batch is still quite large. Each instance involves nearly 360 projection problems, and each projection problem involves about 160 constraints and 350 variables. When we stack constraints into block diagonal form to exploit parallelism, there are about 1,000,000 rows and 2,000,000 columns in the whole matrix. GLinSAT-Sparse-Implicit is the only way that will not report out-of-memory issues when we use GLinSAT. For GLinSAT, the initial estimate of Lipschitz constant is set to $\theta$, the initial estimate of the dual variable is set to zero vector, the numerical precision is set to $10 \epsilon_{\rm{machine}}$, where $\epsilon_{\rm{machine}}$ is the machine epsilon. As to CvxpyLayers and OptNet, neither of them can directly handle such a giant matrix within reasonable time thus we can only use a sequential way instead. The performance of each satisfibility layer is provided in Table {\ref{tab_performance_uc}}. The results of LinSAT layers are not reported in Table {\ref{tab_performance_uc}} since LinSAT layers only support positive linear constraints. All the experiments are conducted on a computer with a Intel(R) Xeon(R) Platinum 8360H CPU and a NVIDIA Tesla A100 GPU with 80GB memory through Pytorch 2.2.

\begingroup
\renewcommand{\arraystretch}{1.1}
\begin{table}[h]
	\small
	\centering
	\caption{Average allocated GPU memory and solution time of different satisfiability layers during batch processing of projection and backpropagation in the training stage of predicting unit states}
	\begin{tabularx}{\textwidth}{M{0.234\textwidth}CCCC}
		\toprule		
		& \multicolumn{2}{c}{GPU Mem./MB} & \multicolumn{2}{c}{Time/s} \\
		\cmidrule(lr){2-3} \cmidrule(lr){4-5} 
		Layer & Proj. & Backprop.  & Proj. & Backprop.  \\
		\midrule
		CvxpyLayers   & \---\---     & \---\---    & 2771     & 684.0   \\
		OptNet  & 33012     & 96.64    & 257.4    & 23.60   \\
		GLinSAT-Sparse-Implicit & \textbf{930.7}    & \textbf{76.86}    & \textbf{26.78}    & \textbf{1.636}    \\
		\bottomrule
	\end{tabularx}%
		\begin{tablenotes}
			\item[a] Note: Statistics of CvxpyLayers and OptNet are based on the first epoch since we cannot obtain a well-trained model in reasonable time.
		\end{tablenotes}
	\label{tab_performance_uc}%
\end{table}%
\endgroup

In the validation stage, we use different values of $\frac{1}{\theta}$ to test their performance. When $\frac{1}{\theta}$ is set to 10\textsuperscript{-3}, we use the double-precision floating-point numbers during projection to avoid potential numerical issues. When $\frac{1}{\theta}$ is set to 0, the projection problem turns into a linear programming problem and we solve it via Gurobi. After we have obtained the outputs of satisfiability layers, we round the outputs of final layers to 0 or 1 and then fix the unit state variables $u_g$ using these rounded outputs. Once we fix all unit state variables to 0 or 1, the original mixed-integer linear programming (MILP) problem turns to a linear programming (LP) problem. We use Gurobi to solve such an LP problem and record whether the problem with fixed unit states is feasible and record the optimal solution if exists. We also compare the optimal objective obtained from fixing variables with the original optimal objective. The corresponding average gaps between them are shown in Table {\ref{tab_result_uc}}.

To illustrate the importance of the satisfiable layer, we additionally substitute the original satisfiability layer with a simple sigmoid activation function in both training and validation stages and report the corresponding result in Table {\ref{tab_result_uc}}. We can easily see that satisfiability layers are essential to produce feasible unit states.


\newpage
\section*{NeurIPS Paper Checklist}

\begin{enumerate}

\item {\bf Claims}
    \item[] Question: Do the main claims made in the abstract and introduction accurately reflect the paper's contributions and scope?
    \item[] Answer: \answerYes{} 
    \item[] Justification: In this paper, we consider making a batch of neural network outputs satisfy bounded and general linear constraints. We present GLinSAT, which is the first general linear satisfiability layer in which all the operations are differentiable and matrix-factorization-free. Experimental results demonstrate the advantages of GLinSAT over existing methods.

\item {\bf Limitations}
    \item[] Question: Does the paper discuss the limitations of the work performed by the authors?
    \item[] Answer: \answerYes{} 
    \item[] Justification: We have discussed our limitations in Appendix {\ref{limitations}}. For variables with one-sided boundary or no explicit boundary, our method cannot be directly used. A possible workaround is to manually calculate the implicit bounds of these variables through domain propagation but we have not implemented such an algorithm in GLinSAT. Also, currently our proposed framework cannot deal with conic constraints.     

\item {\bf Theory Assumptions and Proofs}
    \item[] Question: For each theoretical result, does the paper provide the full set of assumptions and a complete (and correct) proof?
    \item[] Answer: \answerYes{} 
    \item[] Justification: In the derivation process, we always assume the feasible region is non-empty as shown in Sec. {\ref{sec3.1}}. The derivation process of our results can be found in Sec. {\ref{sec3.1}}, Appendix {\ref{appendix.a2}}, {\ref{appendix.a3}} and {\ref{appendix.a4}}.

    \item {\bf Experimental Result Reproducibility}
    \item[] Question: Does the paper fully disclose all the information needed to reproduce the main experimental results of the paper to the extent that it affects the main claims and/or conclusions of the paper (regardless of whether the code and data are provided or not)?
    \item[] Answer: \answerYes{} 
    \item[] Justification: We have detailed our experimental settings in Appendix {\ref{appendix.a5}}, {\ref{appendix.a6}}, {\ref{appendix.a7}} and {\ref{appendix.a8}} and the code released in github can be also be used as a reference.

\item {\bf Open access to data and code}
    \item[] Question: Does the paper provide open access to the data and code, with sufficient instructions to faithfully reproduce the main experimental results, as described in supplemental material?
    \item[] Answer: \answerYes{} 
    \item[] Justification: We provide our codes and instructions on the first three experiments, but currently we are unable to open source the data and code about the unit commitment problem in a real power system due to data security issues and confidentiality agreements.

\item {\bf Experimental Setting/Details}
    \item[] Question: Does the paper specify all the training and test details (e.g., data splits, hyperparameters, how they were chosen, type of optimizer, etc.) necessary to understand the results?
    \item[] Answer: \answerYes{} 
    \item[] Justification: We have detailed our experimental settings in Appendix {\ref{appendix.a5}}, {\ref{appendix.a6}}, {\ref{appendix.a7}} and {\ref{appendix.a8}} and the code released in github can be also be used as a reference.

\item {\bf Experiment Statistical Significance}
    \item[] Question: Does the paper report error bars suitably and correctly defined or other appropriate information about the statistical significance of the experiments?
    \item[] Answer: \answerNo{} 
    \item[] Justification: Error bars are not reported because it would be too computationally expensive.

\item {\bf Experiments Compute Resources}
    \item[] Question: For each experiment, does the paper provide sufficient information on the computer resources (type of compute workers, memory, time of execution) needed to reproduce the experiments?
    \item[] Answer: \answerYes{} 
    \item[] Justification: All the experiments are conducted on a computer with a Intel(R) Xeon(R) Platinum 8360H CPU and a NVIDIA Tesla A100 GPU with 80GB memory through Pytorch 2.2. We also provide the batch processing performance in Table {\ref{tab_performance_tsp}}, Table {\ref{tab_performance_tsp_2}}, Table {\ref{tab_performance_match}}, Table {\ref{tab_performance_port}}, Table {\ref{tab_performance_uc}}.
    
\item {\bf Code Of Ethics}
    \item[] Question: Does the research conducted in the paper conform, in every respect, with the NeurIPS Code of Ethics \url{https://neurips.cc/public/EthicsGuidelines}?
    \item[] Answer: \answerYes{} 
    \item[] Justification: The research conducted in the paper conform with the NeurIPS Code of Ethics.

\item {\bf Broader Impacts}
    \item[] Question: Does the paper discuss both potential positive societal impacts and negative societal impacts of the work performed?
    \item[] Answer: \answerYes{} 
    \item[] Justification: We have discussed potential impacts in Appendix {\ref{broader_impacts}}.
    
\item {\bf Safeguards}
    \item[] Question: Does the paper describe safeguards that have been put in place for responsible release of data or models that have a high risk for misuse (e.g., pretrained language models, image generators, or scraped datasets)?
    \item[] Answer: \answerNA{} 
    \item[] Justification: This paper poses no risk for misusing our proposed method.

\item {\bf Licenses for existing assets}
    \item[] Question: Are the creators or original owners of assets (e.g., code, data, models), used in the paper, properly credited and are the license and terms of use explicitly mentioned and properly respected?
    \item[] Answer: \answerYes{} 
    \item[] Justification: We have properly credited the used Pascal VOC Keypoint dataset {\cite{everingham2010pascal}} with Berkeley annotations {\cite{bourdev2009poselets}} under the unfiltered setting {\cite{wang2023linsatnet, rolinek2020deep}}. We also cite the TSP dataset and portfolio allocation dataset provided in Ref. {\cite{wang2023linsatnet}}.

\item {\bf New Assets}
    \item[] Question: Are new assets introduced in the paper well documented and is the documentation provided alongside the assets?
    \item[] Answer: \answerNA{} 
    \item[] Justification: This paper does not release new assets.

\item {\bf Crowdsourcing and Research with Human Subjects}
    \item[] Question: For crowdsourcing experiments and research with human subjects, does the paper include the full text of instructions given to participants and screenshots, if applicable, as well as details about compensation (if any)? 
    \item[] Answer: \answerNA{} 
    \item[] Justification: This paper does not involve crowdsourcing nor research with human subjects.

\item {\bf Institutional Review Board (IRB) Approvals or Equivalent for Research with Human Subjects}
    \item[] Question: Does the paper describe potential risks incurred by study participants, whether such risks were disclosed to the subjects, and whether Institutional Review Board (IRB) approvals (or an equivalent approval/review based on the requirements of your country or institution) were obtained?
    \item[] Answer: \answerNA{} 
    \item[] Justification: This paper does not involve crowdsourcing nor research with human subjects.

\end{enumerate}

\end{document}